\documentclass[accepted]{uai2023} % after acceptance, for a revised
\usepackage[american]{babel}

\usepackage[round]{natbib} % has a nice set of citation styles and commands
\usepackage{mathtools} % amsmath with fixes and additions
\usepackage{siunitx} % for proper typesetting of numbers and units
\usepackage{booktabs} % commands to create good-looking tables
\usepackage{tikz} % nice language for creating drawings and diagrams

\usepackage{xspace}
\usepackage{multirow}
\usepackage{caption}
\usepackage{subcaption}
\RequirePackage{algorithm}
\RequirePackage{algorithmic}
\usepackage{enumitem}
\usepackage{xspace}

\usepackage{amssymb}% http://ctan.org/pkg/amssymb
\usepackage{pifont}% http://ctan.org/pkg/pifont
\newcommand{\cmark}{\ding{51}}%
\newcommand{\xmark}{\ding{55}}%

\usepackage{mathtools}      % Mathematical tools to use with amsmath
\usepackage{amssymb}
\usepackage{amsthm}         % Basic mathematical environments for proofs etc.
\usepackage{mathdots}       % dots commands for matrices 
\usepackage{amsfonts}       % blackboard math symbols
\usepackage{nicefrac}       % compact symbols for 1/2, etc.
\usepackage{physics}        % typesetting equations for physics simpler
\usepackage{bm}             % Access bold symbols in maths mode
\usepackage{centernot}      % The package provides \centernot
\usepackage{thmtools}       % Extensions to theorem environments

\DeclareMathOperator*{\argmax}       {arg\,max}

\DeclareMathOperator*{\Pbb}          {\mathbb{P}}

\makeatletter
\DeclareRobustCommand\onedot{\futurelet\@let@token\@onedot}
\def\@onedot{\ifx\@let@token.\else.\null\fi\xspace}
\def\eg{\emph{e.g}\onedot} 
\def\ie{\emph{i.e}\onedot} 
\makeatother

\def\ddefloop#1{\ifx\ddefloop#1\else\ddef{#1}\expandafter\ddefloop\fi}

\def\ddef#1{\expandafter\def\csname #1bb\endcsname{\ensuremath{\mathbb{#1}}}}
\ddefloop ABCDEFGHIJKLMNOPQRSTUVWXYZ\ddefloop

\def\ddef#1{\expandafter\def\csname #1cal\endcsname{\ensuremath{\mathcal{#1}}}}
\ddefloop ABCDEFGHIJKLMNOPQRSTUVWXYZ\ddefloop

\def\ddef#1{\expandafter\def\csname #1mat\endcsname{\ensuremath{\mathbf{#1}}}}
\ddefloop ABCDEFGHIJKLMNOPQRSTUVWXYZ\ddefloop

\def\ddef#1{\expandafter\def\csname #1matsf\endcsname{\ensuremath{\mathsf{#1}}}}
\ddefloop ABCDEFGHIJKLMNOPQRSTUVWXYZ\ddefloop

\def\ddef#1{\expandafter\def\csname #1vec\endcsname{\ensuremath{\mathbf{#1}}}}
\ddefloop abcdefghijklmnopqrstuvwxyz\ddefloop

\newcommand{\abstain}{{\ensuremath{\oslash}}\xspace}
\newcommand{\pval}{{\ensuremath{pv}}\xspace}

\newtheorem{theorem}{Theorem}

\title{Towards Better Certified Segmentation via Diffusion Models}

\author[1]{Othmane Laousy}
\author[2]{Alexandre Araujo}
\author[3]{Guillaume Chassagnon}
\author[3]{Marie-Pierre Revel}
\author[2]{Siddharth Garg}
\author[2]{Farshad Khorrami}
\author[1]{Maria Vakalopoulou}

\affil[1]{%
    MICS, CentraleSupélec, Paris-Saclay University, Inria Saclay, France
}
\affil[2]{%
    New York University, NY, USA
}
\affil[3]{%
    Paris Cité University, France
}

\begin{document}
\maketitle

\begin{abstract}
The robustness of image segmentation has been an important research topic in the past few years as segmentation models have reached production-level accuracy.
However, like classification models, segmentation models can be vulnerable to adversarial perturbations, which hinders their use in critical-decision systems like healthcare or autonomous driving.
Recently, randomized smoothing has been proposed to certify segmentation predictions by adding Gaussian noise to the input to obtain theoretical guarantees.
However, this method exhibits a trade-off between the amount of added noise and the level of certification achieved.
In this paper, we address the problem of certifying segmentation prediction using a combination of randomized smoothing and diffusion models. 
Our experiments show that combining randomized smoothing and diffusion models significantly improves certified robustness, with results indicating a mean improvement of 21 points in accuracy compared to previous state-of-the-art methods on Pascal-Context and Cityscapes public datasets. Our method is independent of the selected segmentation model and does not need any additional specialized training procedure.
\end{abstract}

\section{Introduction}\label{sec:intro}

Neural networks have been known to be vulnerable to adversarial perturbations~\citep{szegedy2013intriguing,madry2017towards,goodfellow2014explaining,carlini2017towards}, \ie, imperceptible variations of natural examples, crafted to deliberately mislead the models.
In recent years, significant efforts have been made to develop certified defenses that guarantee a specified level of robustness against adversarial inputs within a certain radius.
(\eg, 1-Lipschitz Networks~\citep{trockman2021orthogonalizing,meunier2022dynamical,araujo2023a}, bound propagation~\citep{gowal2018effectiveness,huang2021training}, randomized smoothing~\citep{li2019certified,cohen2019certified,salman2019provably}).
Although most defenses focus on classification tasks, in this paper, we focus on certifying segmentation models and argue that certified segmentation is an even more pressing issue  as these models are already used in critical systems such as healthcare and autonomous vehicles.

Randomized smoothing has emerged as the leading technique for certified robustness due to its scalability and  model-agnostic properties.
It consists in applying a convolution between a base classifier and a Gaussian distribution, enabling the method to handle large input sizes (\eg, ImageNet, Pascal-Context, Cityscapes), while providing state-of-the-art certified accuracy.
However, this technique exhibits a trade-off between adding enough noise for certification and preserving the input's semantic information for accurate predictions.
In fact, several impossibility results from an information-theory perspective have been introduced ~\citep{kumar2020curse,blum2020random,yang2020randomized} and inherently limit randomized smoothing from providing large certified radii.
Nevertheless, recent works, both theoretical~\citep{ettedgui2022towards,mohapatra2020higher} and empirical~\citep{salman2020denoised,carlini2023certified}, have explored potential solutions to this trade-off.
To address the issue of removed information due to noise injection, several works, in the context of classification tasks, have proposed methods to {\em denoise} the input after the noise injection step~\citep{salman2020denoised,carlini2023certified}.
While \cite{salman2020denoised} trained their own denoiser models on Gaussian noise for the specific task of certified robustness, \cite{carlini2023certified} extended the work of~\cite{salman2020denoised} by using off-the-shelf {\em Denoising Diffusion Probabilistic Models}~\citep{sohl2015deep,ho2020denoising,nichol2021improved}, a form of generative models that takes a random Gaussian noise and generates a real-world image.

\begin{figure*}[th]
  \centering
  \hfill
  \begin{subfigure}[h]{0.24\textwidth}
    \centering
    \includegraphics[width=\textwidth]{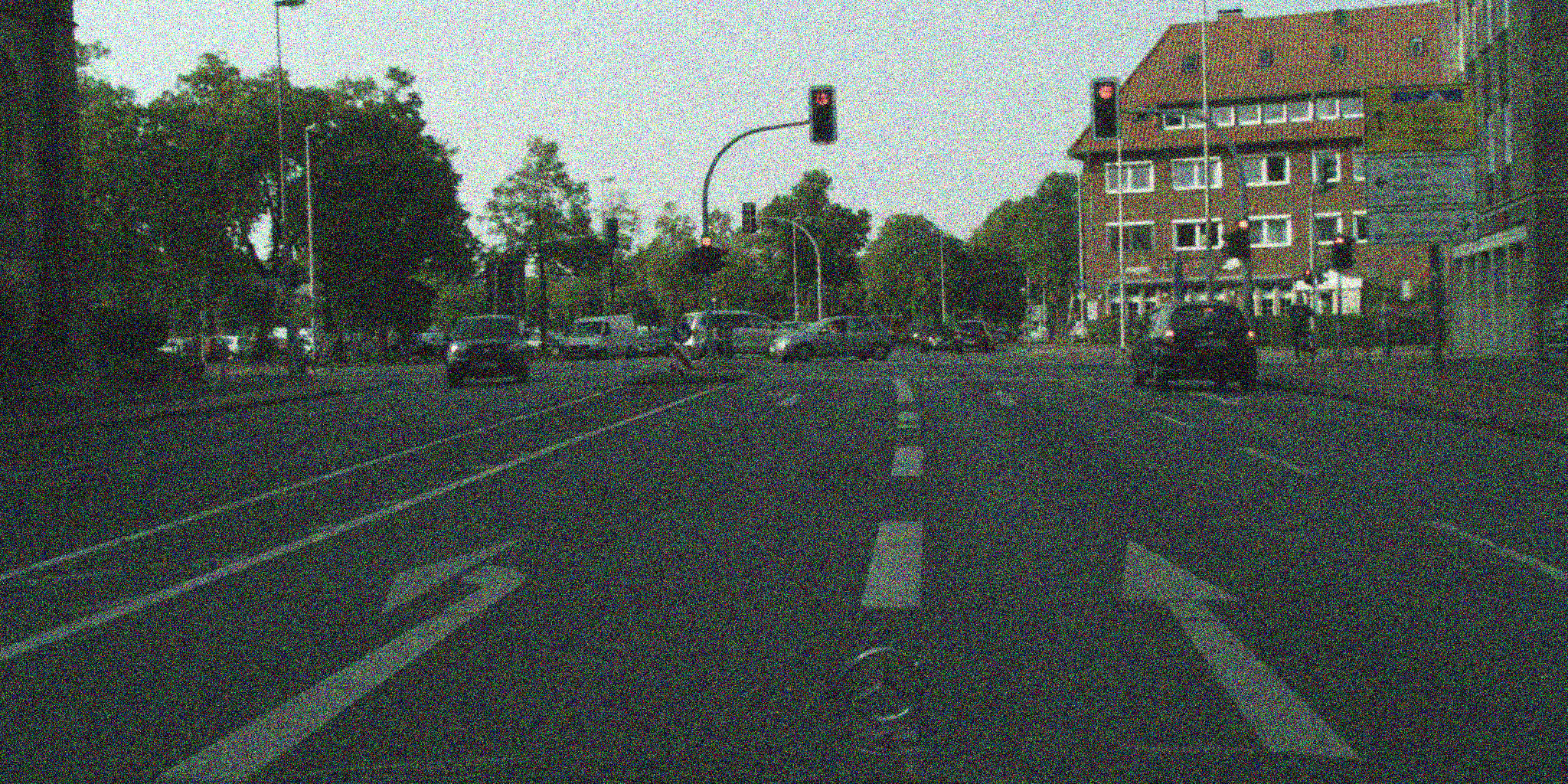} \\[0.2cm]
    \includegraphics[width=\textwidth]{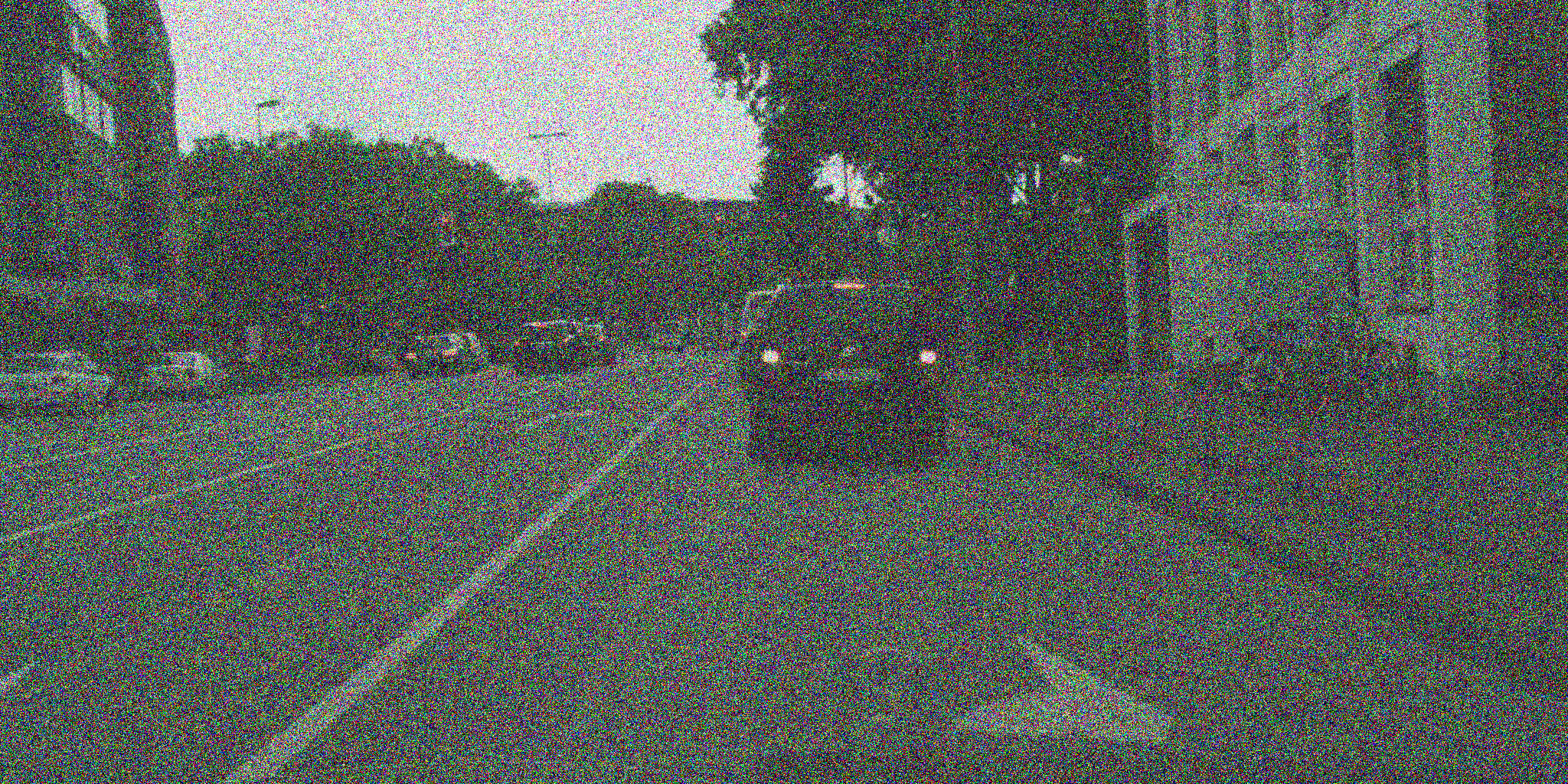} \\[0.2cm]
    \includegraphics[width=\textwidth]{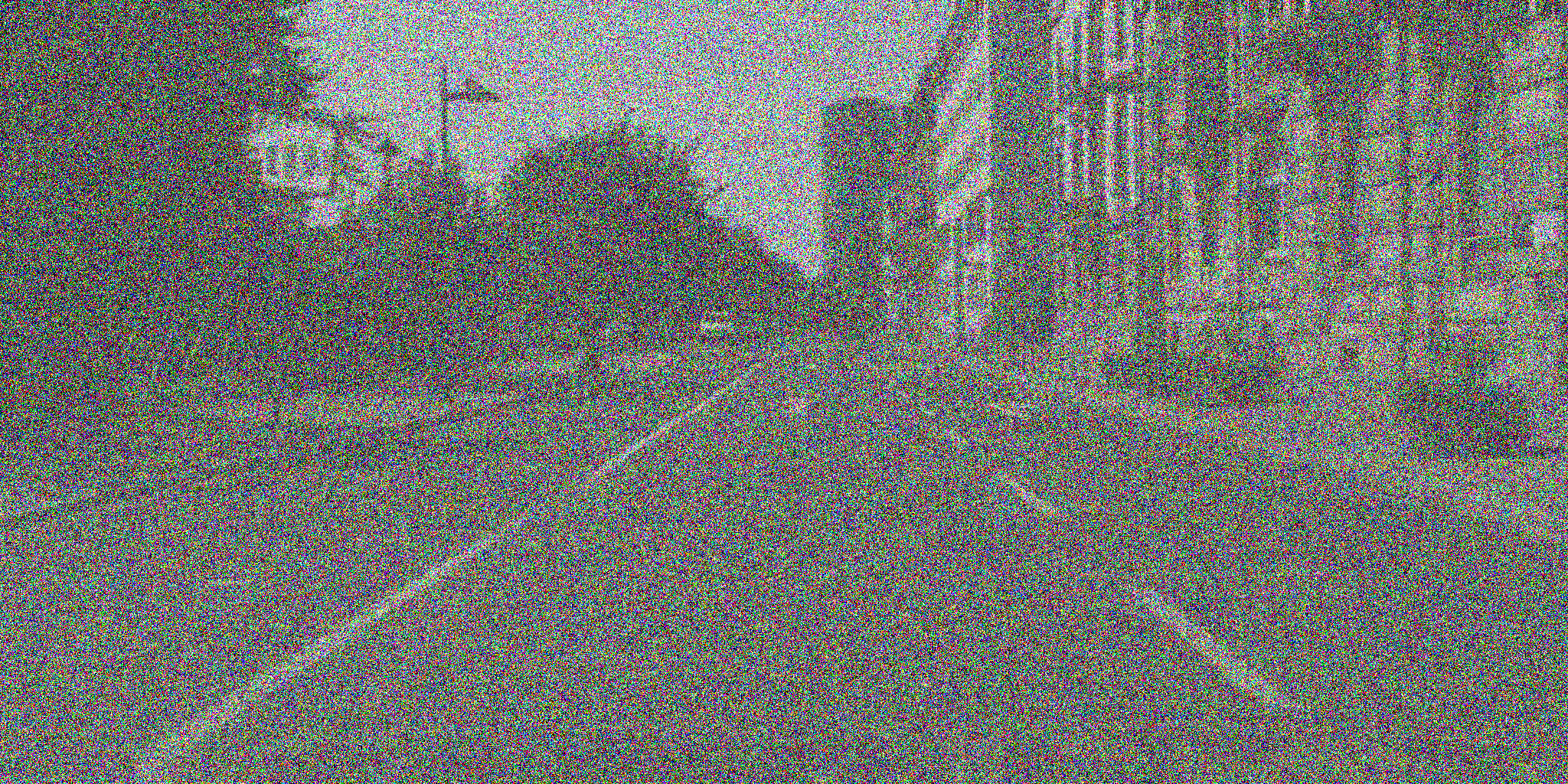}
    \label{subfig:image-a}
    \caption{Image with noise}
  \end{subfigure}
  \hfill
  \begin{subfigure}[h]{0.24\textwidth}
    \centering
    \includegraphics[width=\textwidth]{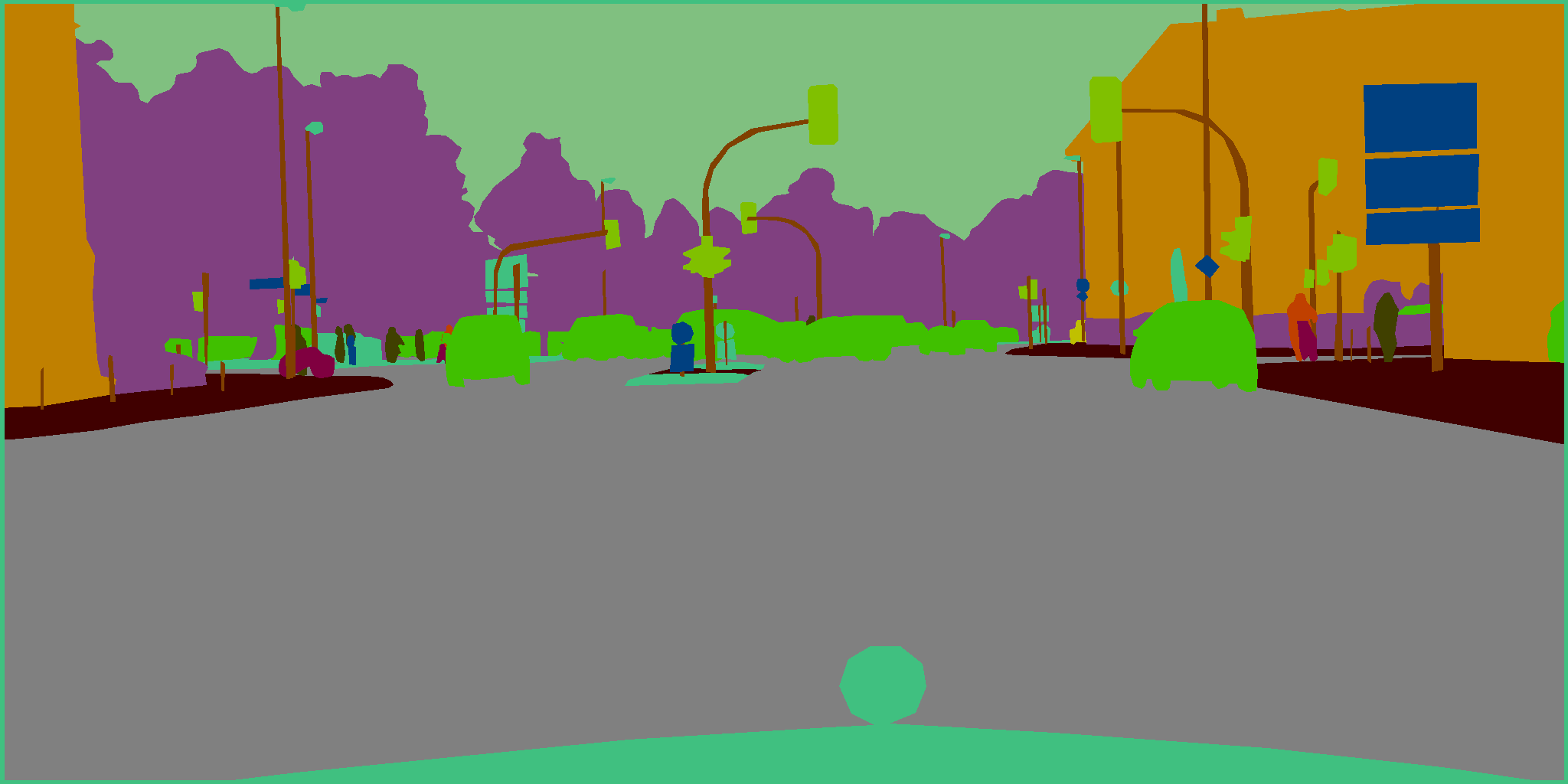} \\[0.2cm]
    \includegraphics[width=\textwidth]{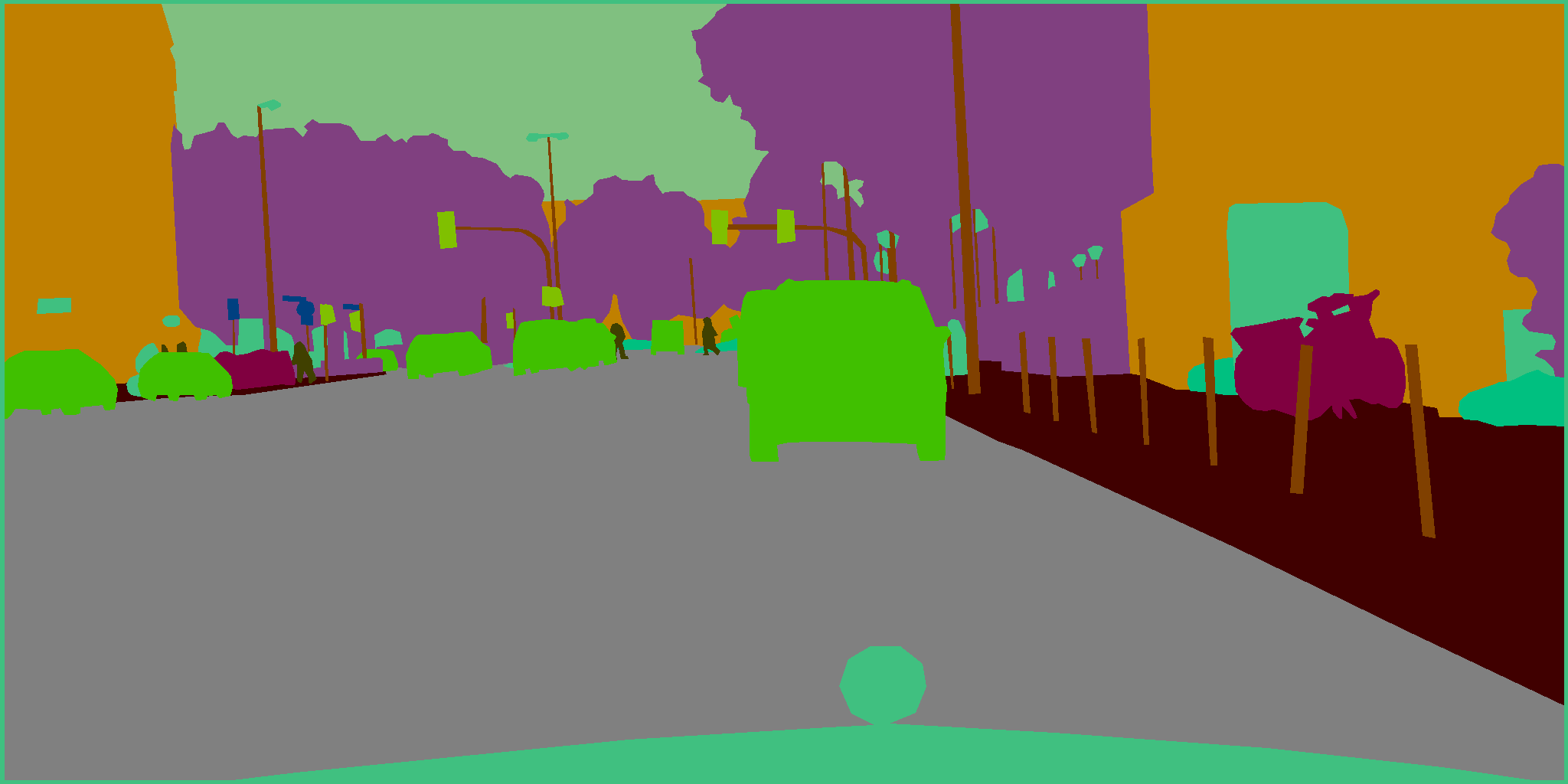} \\[0.2cm]
    \includegraphics[width=\textwidth]{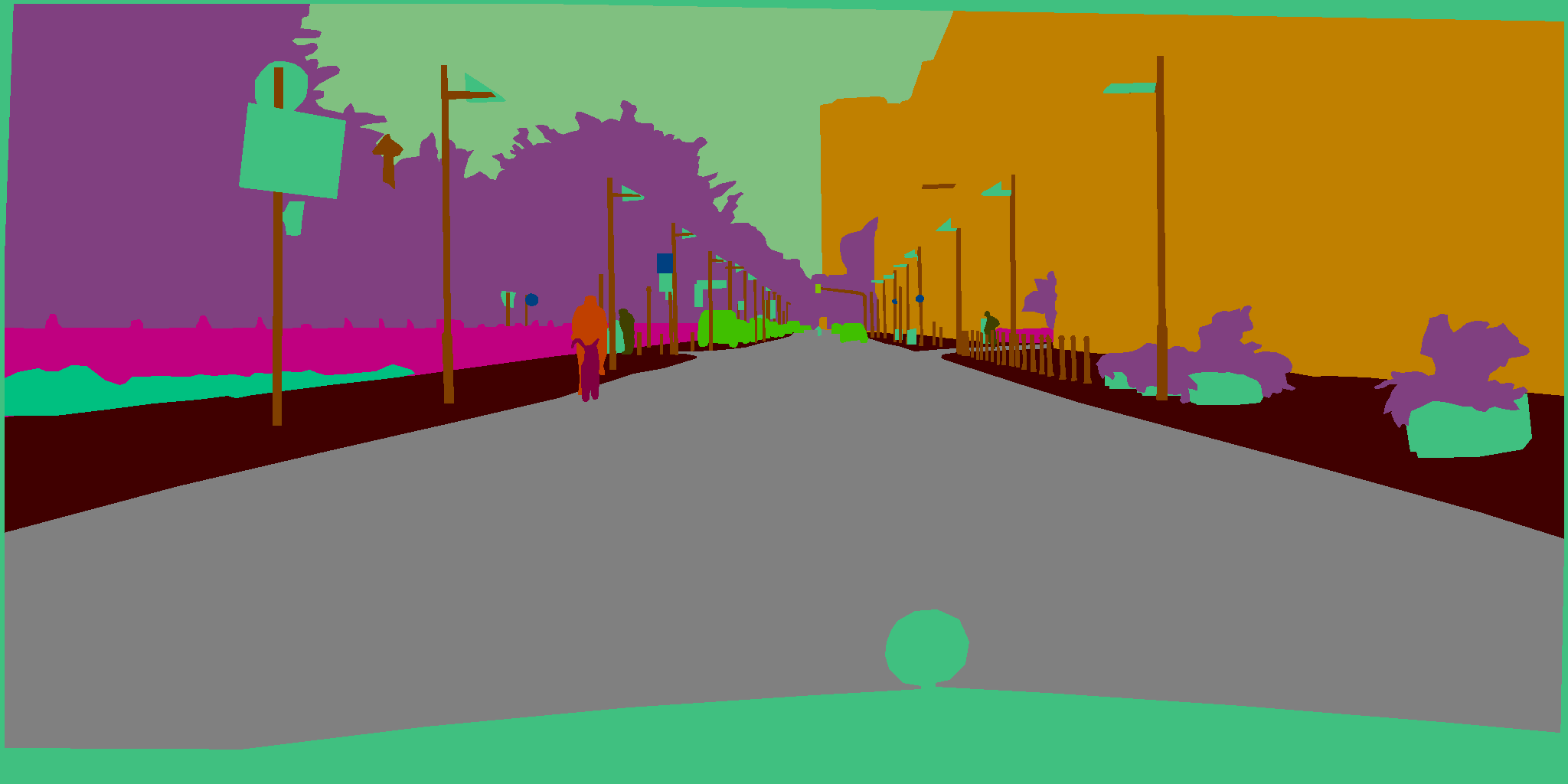}
    \label{subfig:image-b}
    \caption{Ground truth segmentation}
  \end{subfigure}
  \hfill
  \begin{subfigure}[h]{0.24\textwidth}
    \centering
    \includegraphics[width=\textwidth]{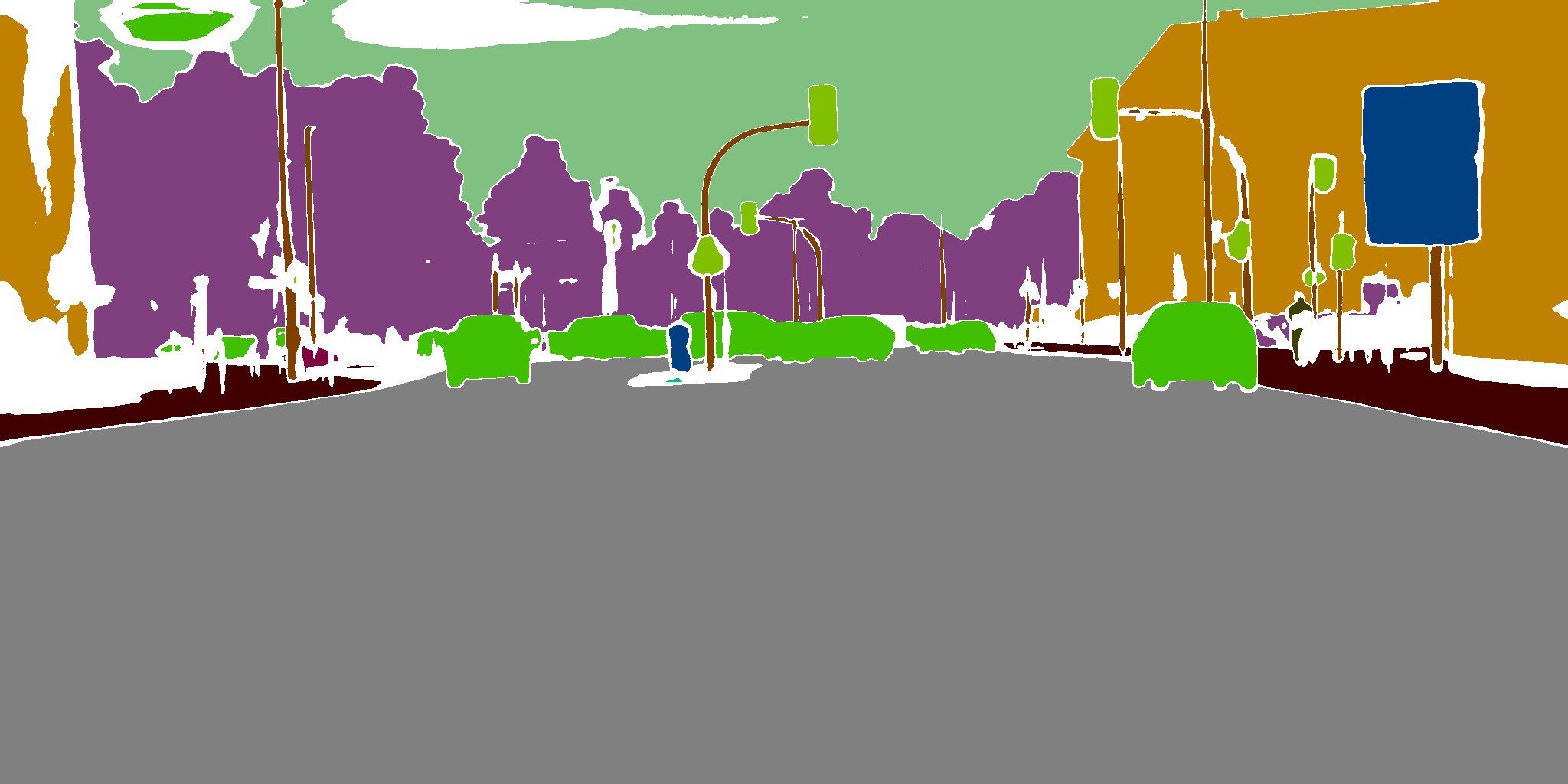} \\[0.2cm]
    \includegraphics[width=\textwidth]{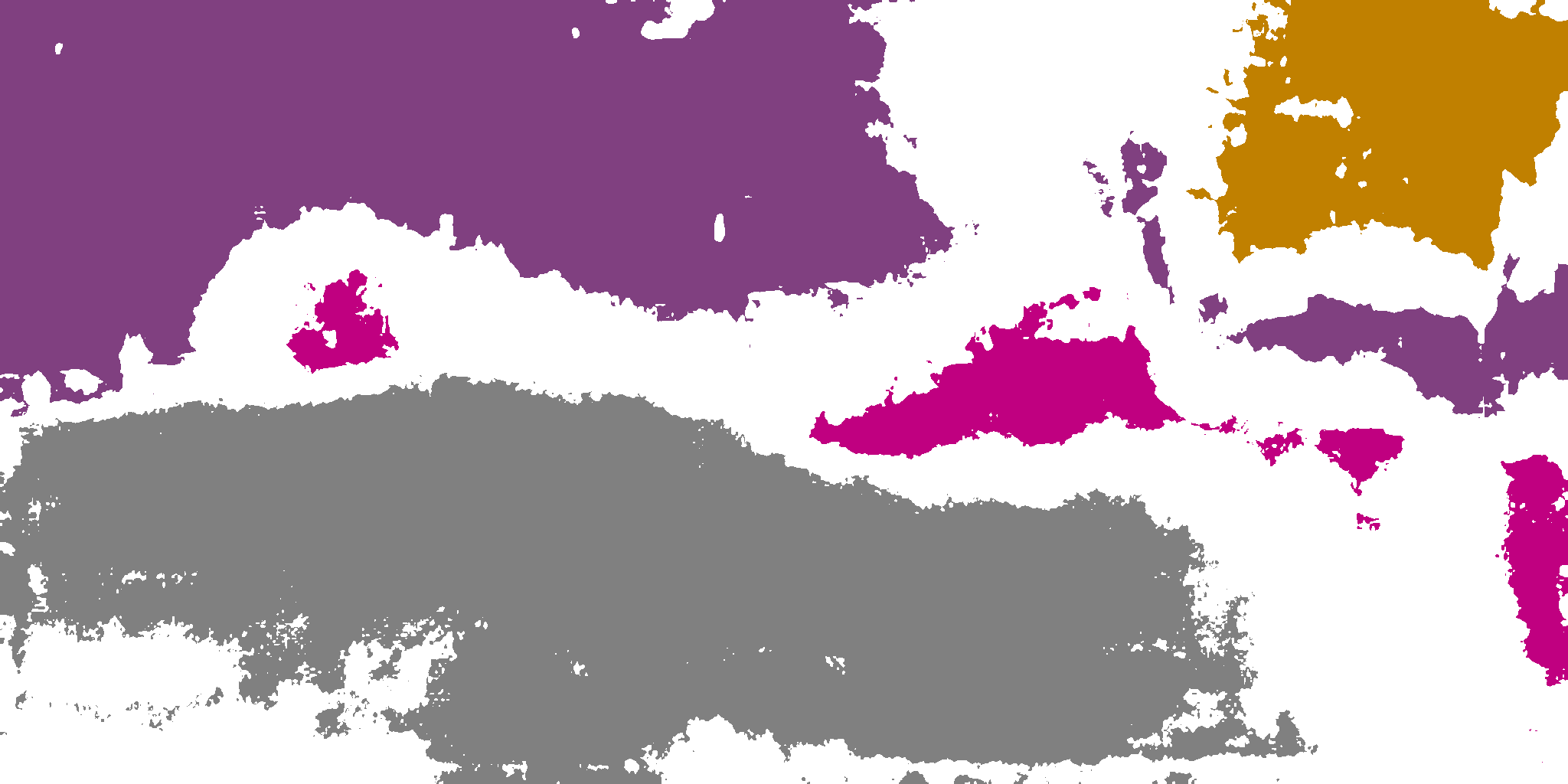} \\[0.2cm]
    \includegraphics[width=\textwidth]{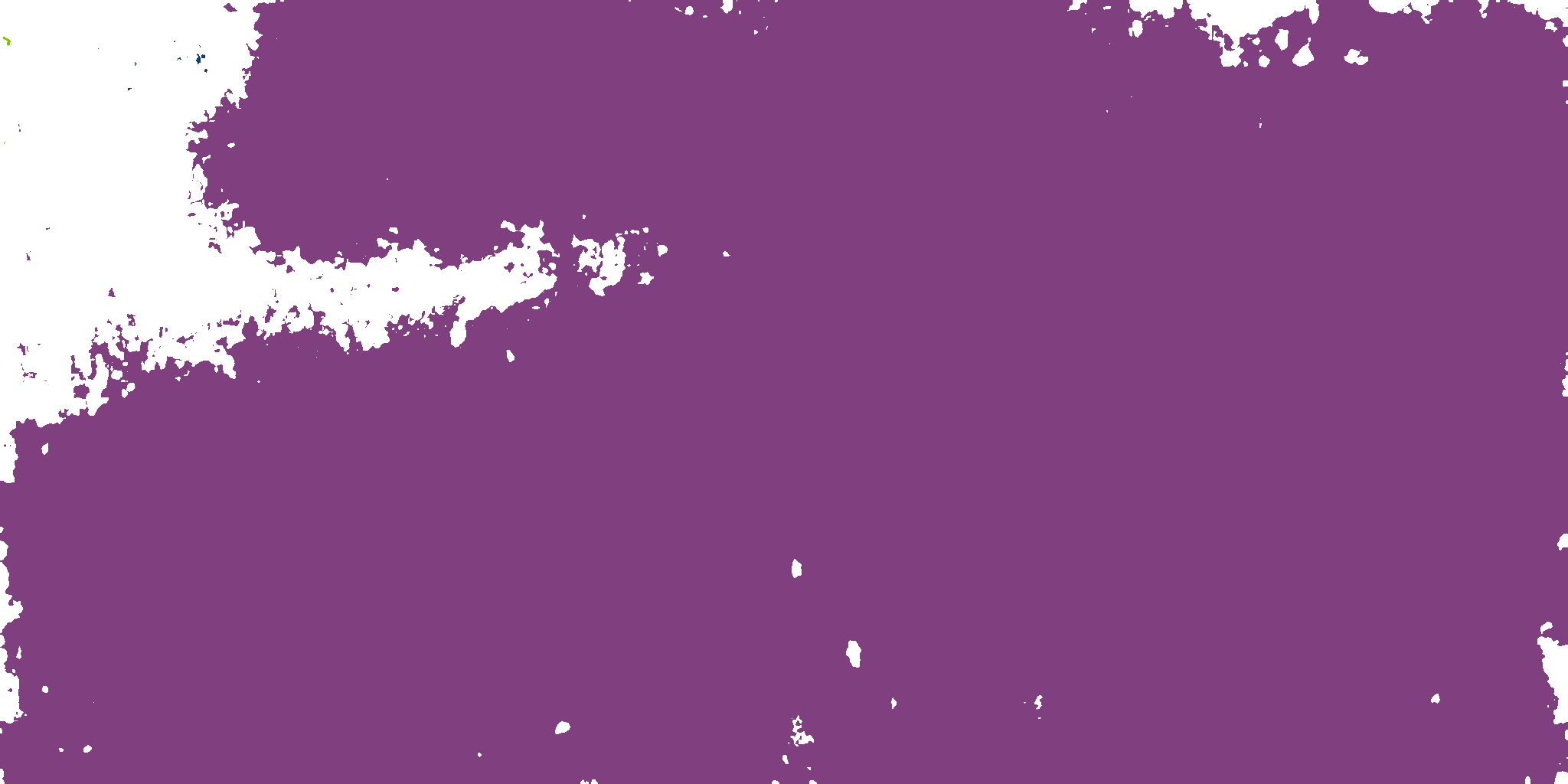}
    \label{subfig:image-c}
    \caption{\textsc{SegCertify} segmentation}
  \end{subfigure}
  \hfill
  \begin{subfigure}[h]{0.24\textwidth}
    \centering
    \includegraphics[width=\textwidth]{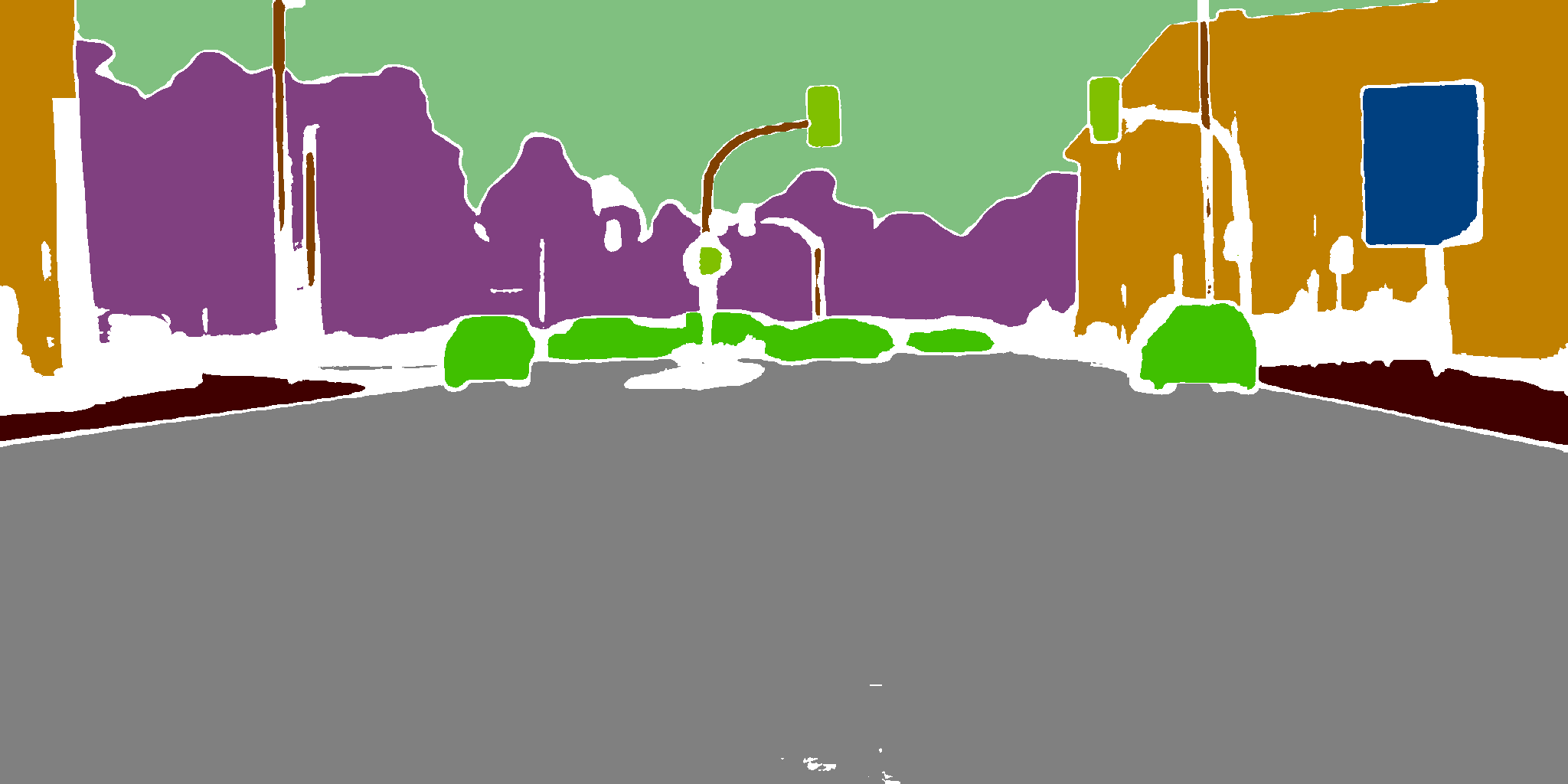} \\[0.2cm]
    \includegraphics[width=\textwidth]{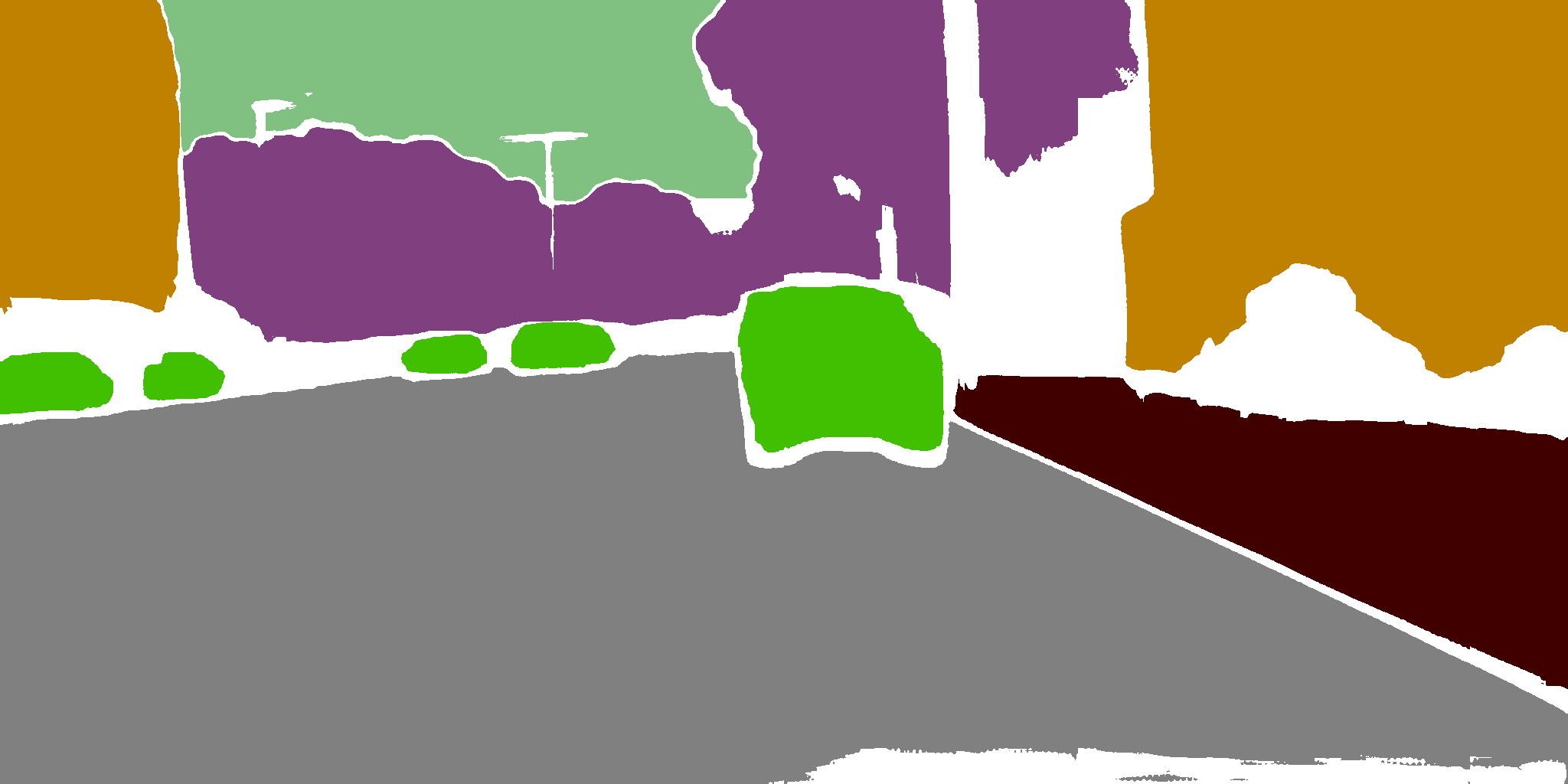} \\[0.2cm]
    \includegraphics[width=\textwidth]{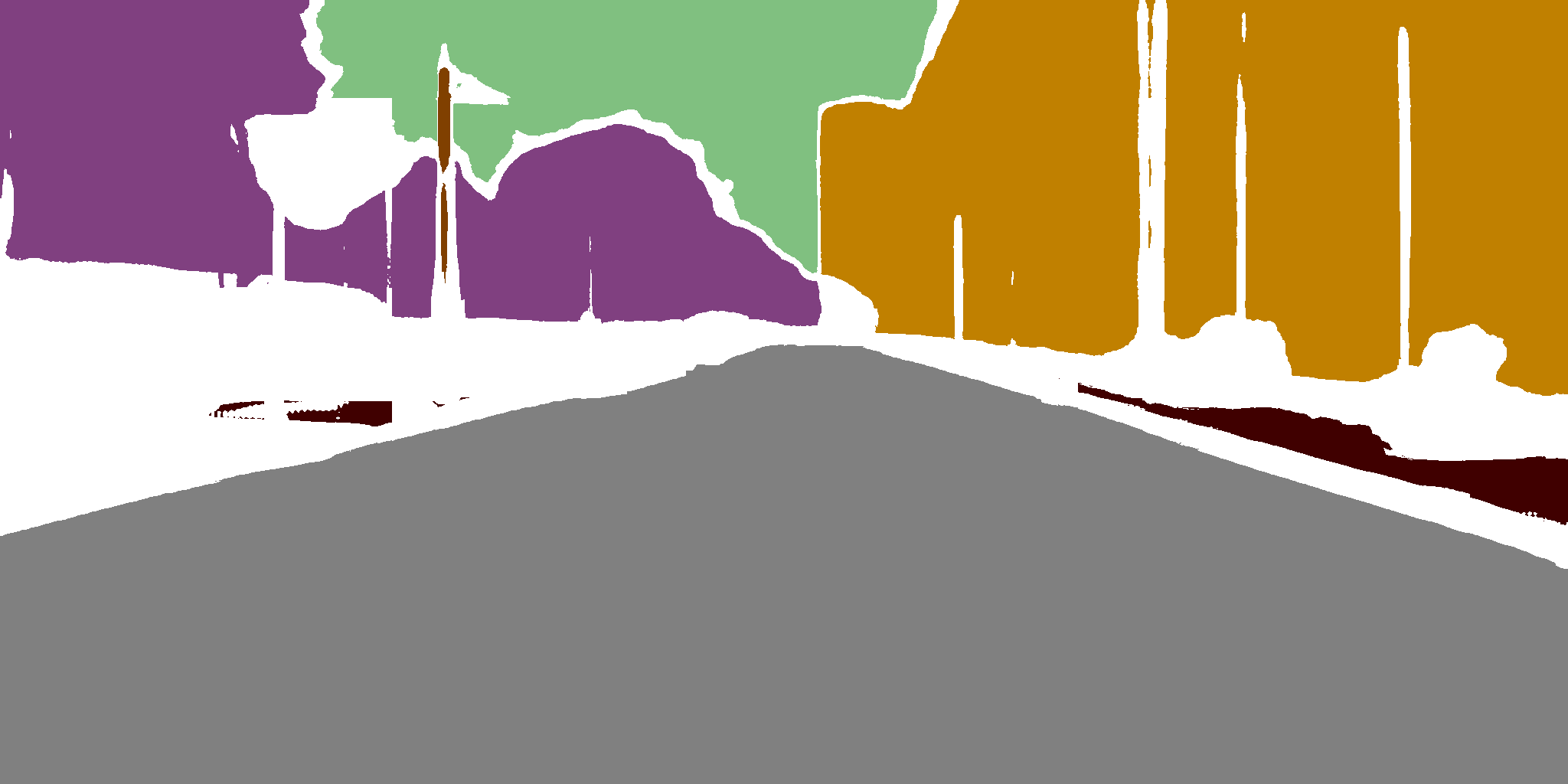}
    \label{subfig:image-d}
    \caption{Our segmentation}
  \end{subfigure}
  \caption{Examples of our approach (\textsc{DenoiseCertify}) compared to  \textsc{SegCertify} proposed by \cite{fischer2021scalable} on the Cityscapes dataset. From left to right: (a) the initial image with added noise, (b) the ground truth segmentation, (c) the abstained segmentation obtained with \textsc{SegCertify}, (d) the abstained segmentation obtained with \textsc{DenoiseCertify} (ours). Each row corresponds to a noise level, from top to bottom: $\sigma = 0.25, 0.5$ and $1.0$.}
  \label{figure:cityscapes}
\end{figure*}

In this paper, we build upon previous work on certified robustness to improve certified segmentation extending the work of~\cite{fischer2021scalable} and~\cite{carlini2023certified}.
We present a comprehensive set of experiments on PASCAL-Context~\citep{mottaghi2014role} and Cityscapes~\citep{cordts2016cityscapes} datasets and successfully achieve state-of-the-art results on certified robustness for segmentation tasks. 
Our results show that combining randomized smoothing and diffusion models significantly improves certified robustness, with a mean increase of 21 points in accuracy and 14 points in mIoU when compared to previous methods.
Our main contributions are summarized as follows:

\begin{itemize}
    \item First, we build upon the work of \cite{fischer2021scalable} and \cite{carlini2023certified} and propose for the first time, a certified segmentation approach leveraging diffusion models.
    Through a series of experiments, we demonstrate that incorporating a denoiser in conjunction with a segmentation model that has been {\em trained with noise injection} presents certain trade-offs in the certified accuracy achieved, depending on the variance of the noise.
     \item Second, we further improve certified accuracy by combining off-the-shelf diffusion and state-of-the-art segmentation models allowing us to reach state-of-the-art results for certified segmentation.
    \item Third, we propose an in-depth analysis through a series of experiments on the use of noise during training as well as the generalization of denoising diffusion models with respect to image resolution and data distribution.
\end{itemize}

\section{Related Work}

\textbf{Adversarial Attacks \& Certified Defenses.}
Since the discovery of adversarial examples~\citep{szegedy2013intriguing}, a wealth of work focused on devising attacks~\citep{goodfellow2014explaining,kurakin2018adversarial,carlini2017towards,croce2020reliable,croce2021mind} and defenses~\citep{goodfellow2014explaining,madry2017towards,pinot2019theoretical,araujo2020advocating,araujo2021lipschitz}, leading to an ongoing back-and-forth battle.
Most of these defenses relied on smoothing the local neighborhood around each point, resulting in very small gradients on which attacks were based.
However, it has become apparent that many of the empirical defenses that have been created could be circumvented with stronger attacks~\citep{athalye2018obfuscated}. 

This false sense of security and the persistent cat-and-mouse game called for {\em certified defenses} that provide provable robustness guarantees.
In recent years, mainly two types of certified defenses have been proposed.
The first approach provides robustness guarantees based on the Lipschitz constant of the networks and their margin (\ie, the difference between the highest and second highest logits).
This connection was introduced by~\cite{tsuzuku2018lipschitz} and opened an important research direction in the design and training of 1-Lipschitz neural networks~\citep{miyato2018spectral,farnia2018generalizable,li2019preventing,trockman2021orthogonalizing,singla2021skew,yu2022constructing,meunier2022dynamical,prach2022almost,xu2022lot,araujo2023a}.
Although this approach offers fast certificate computation, it suffers from important drawbacks.
Indeed, due to the strict constraint on the networks and reduced expressivity, 1-Lipschitz neural networks offer a reduced natural and certified accuracy and do not scale to large datasets (\eg, ImageNet, Pascal-Context, Cityscapes).
On the other hand, a second approach called Randomized Smoothing leverages randomization.
This method, introduced by~\cite{lecuyer2019certified} and further improved by~\cite{li2019certified,cohen2019certified} and \cite{salman2019provably}, consists in convolving the function with a Gaussian probability distribution during the inference phase.
The desirable property of a smooth classifier is ensuring that the prediction is constant within an $\ell_2$ ball around any input.

\textbf{Diffusion models.}
Diffusion probabilistic models have been introduced by~\cite{sohl2015deep}, and further refined by~\cite{ho2020denoising} and \cite{nichol2021improved}.
The goal was to design a generative Markov chain that transforms a known distribution (\eg, Gaussian) into a target (data) distribution using a diffusion process.
However, instead of using a Markov chain to evaluate the model, they defined the probabilistic model as the endpoint of the Markov chain.
Subsequently, this methodology was refined and applied for producing high-quality samples, such as images, as demonstrated by~\cite{ho2020denoising} and \cite{nichol2021improved}.
The results indicated that this type of model can generate better images in comparison to other methods and also demonstrated a connection with denoising.
Recently, diffusion probabilistic models have been applied successfully in the context of certified robustness for classification tasks where a diffusion model is used as a first step to denoise inputs for randomized smoothing~\citep{carlini2023certified}.

\textbf{Certified Segmentation.}
Deep neural networks trained for segmentation tasks have been shown to be vulnerable to adversarial attacks~\citep{xie2017adversarial,arnab2018robustness,xiang2019generating,he2019non,kang2020adversarial}.
In this context, \cite{fischer2021scalable} use the work of~\cite{cohen2019certified} and propose a method to certify segmentation with randomized smoothing for norm-bounded perturbations.
Other lines of work investigate certified robustness for structured outputs, for example, \cite{kumar2021center} proposed a procedure based on randomized smoothing to find the minimum enclosing ball in the output space and \cite{yatsura2022certified} introduced a method called \emph{demasked smoothing} to defend against adversarial patch attacks for semantic segmentation tasks.

In this paper, we build upon the work of \cite{fischer2021scalable} and \cite{carlini2023certified} and introduce, for the first time, a randomized smoothing approach with a denoising step in the context of certified semantic segmentation.

\begin{figure*}
  \begin{minipage}[t]{0.48\textwidth}
  \begin{algorithm}[H]
     \caption{Predict \& Certify by~\cite{fischer2021scalable}}
     \label{alg:segcertify}
  \begin{algorithmic}[1]
     \STATE \textbf{function} \textsc{SegCertify}($g$, $\sigma$, $x$, $n$, $n_{0}$, $\delta$, $\alpha$)
     \STATE \quad $\text{\texttt{cnts}}^{0}_{1}, \dots, \text{\texttt{cnts}}^{0}_{N}$ $\leftarrow$ \textsc{Sample}($g$, $x$, $n_{0}$, $\sigma$)
     \STATE \quad $\text{\texttt{cnts}}_{1}, \dots,  \text{\texttt{cnts}}_{N}$ $\leftarrow$ \textsc{Sample}($g$, $x$, $n$, $\sigma$)
     \STATE \quad\textbf{for} $i \leftarrow \{1, \dots, N\}$:
     \STATE \quad\quad $\hat{c}_{i} \leftarrow$ top index in $\text{\texttt{cnts}}^{0}_{i}$
     \STATE \quad\quad $n_{i} \leftarrow \text{\texttt{cnts}}_{i}[\hat{c}_{i}]$
     \STATE \quad\quad $\pval_{i} \leftarrow $\textsc{BinPValue}($n_i$, $n$, $\leq$, $\delta$)
     \STATE \quad $r_{1}, \dots, r_{N} \leftarrow $ \textsc{FwerControl}($\alpha$, $\pval_{1}$, $\dots$, $\pval_{N}$)
     \STATE \quad\textbf{for} $i \leftarrow \{1, \dots, N\}$:
     \STATE \quad\quad \textbf{if} $\lnot r_{i}$: $\hat{c}_{i} \leftarrow \abstain$
     \STATE \quad $R \leftarrow \sigma \Phi^{-1}(\delta)$
     \STATE \quad \textbf{return} $\hat{c}_{1}, \dots, \hat{c}_{N}$, $R$
  \end{algorithmic}
  \end{algorithm}
  \end{minipage}
  \hfill
  \begin{minipage}[t]{0.48\textwidth}
  \begin{algorithm}[H]
   \caption{Sample Function}
   \label{algo:denoise}
   \begin{algorithmic}[1]
      \STATE \textbf{function} \textsc{Sample}($g$, $x$, $n$, $\sigma$)
      \STATE \quad $\texttt{cnts} \gets []$
      \STATE \quad \textbf{for} $0$ to $n-1$ \textbf{do}
      \STATE \quad \quad $t^\star, \beta_{t^\star} \gets \texttt{computeTimestep}(\sigma)$
      \STATE \quad \quad $x_{t^\star} \gets \sqrt{\beta_{t^\star}} (x + \Ncal(0, \sigma^2 \mathbf{I}))$
      \STATE \quad \quad $y \gets g(\texttt{denoise}(x_{t^\star}; t^\star))$
      \STATE \quad \quad $\texttt{cnts}_y \gets \texttt{cnts}_y + 1$
      \STATE \quad \textbf{return} $\texttt{cnts}$
      \STATE
      \STATE \textbf{function} \textsc{computeTimestep}$(\sigma)$
      \STATE \quad $t^\star \gets \text{find } t \quad \text{s.t.} \quad \frac{1- \beta_t}{ \beta_t} = \sigma^2$
      \STATE \quad \textbf{return} $t^\star, \beta_{t^\star}$
   \end{algorithmic}
  \end{algorithm}
  \end{minipage}
\end{figure*}

\section{Background}

In this section, we review the necessary background on randomized smoothing and on certified segmentation.

\subsection{Adversarial Attacks \& Randomized smoothing for classification}

We first introduce adversarial attacks and randomized smoothing in the setting they have been introduced, \ie, for a classification task. We will generalize it to the segmentation task in the next paragraph.

Let $\Xcal \subset \Rbb^d$ and $\Ycal = \{ 1, \dots, K \}$ be the input space and target space respectively with $K$ denoting the number of classes.
Let us denote a classifier $f: \Xcal \rightarrow \Ycal$ (\eg, a neural network)
such that
for a given couple input-label $(x, y) \in \Xcal \times \Ycal$, we say the classifier $f$ correctly classifies $x$ if: $f(x) = y$.
An adversarial attack is a small norm-bounded perturbation $\delta \in \Rbb^d$ with $\norm{\delta}_2 \leq \epsilon$ such that:
\begin{equation}
    f(x + \delta) \neq y.
\end{equation}

Randomized smoothing, introduced in~\cite{cohen2019certified}, considers a {\em smooth} version of the classifier $f$, such that:
\begin{equation} \label{eq:g_classifier}
  g(x) = \argmax_c \Pbb_{\eta \sim \Ncal(0, \sigma^2 \mathbf{I}) } \left[ f(x + \eta) = c \right] \,.
\end{equation}
To compute the probability in Equation~\ref{eq:g_classifier}, \cite{cohen2019certified} proposed a Monte-Carlo approach where the prediction is computed from a small number of samples, \ie, $n_0$, with a majority vote and a lower-bound on the certified radius computed with a higher number of samples, \ie, $n$.
A benefit of using the smooth classifier $g$ is obtaining a certified radius of robustness for each data point, thus determining a certified level of accuracy within a specified attack `budget' $\epsilon$.
More formally, \citeauthor{cohen2019certified} introduced the following theorem:
\begin{theorem}[From \cite{cohen2019certified}] \label{thm:cohen2019}
  Suppose $y \in \Ycal$, let 
  \begin{equation}
    p_y = \Pbb_{\eta \sim \Ncal(0, \sigma^2 \Imat) } \left[ f(x + \eta) = y \right]
  \end{equation}
  and let $\underline{p_y}$ the lower bound of $p_y$ computed via Monte-Carlo sampling.
  Let $\overline{p_{\lnot y}} = 1 - \underline{p_y}$, then, if
  \begin{equation}
      \Pbb_{\eta}\left[ f(x + \eta)=y \right] \geq 
      \underline{p_y} \geq \overline{p_{\lnot y}} \geq \max_{c \neq y}
      \Pbb_{\eta} \left[ f(x + \eta) = c \right],
  \end{equation}
  then $g(x + \delta) = y$ for all $\delta$ satisfying $\norm{\delta}_2 \leq R$
  with $R := \sigma\Phi^{-1}(\underline{p_y})$ and $\Phi$ is the cumulative distribution function of the standard Gaussian distribution.
\end{theorem}

To properly approximate the probability $p_y$ with a confidence interval, \cite{cohen2019certified} proposed a procedure which samples $n$ realizations of $\eta \sim \Ncal(0, \sigma^2 \Imat)$ and computes $f(x + \eta)$.
From these $n$ realizations, a vector of counts for each class in $\Ycal$ is computed and these counts are then used to estimate the probability $p_y$ and the radius $R$ with confidence $1 - \alpha$ with $\alpha \in [0, 1]$.
If the confidence level is not reached (for example, the number of samples is not enough), the procedure will abstain.

\subsection{Certified Segmentation via Randomized Smoothing}

\cite{fischer2021scalable} extended the work of~\cite{cohen2019certified} for segmentation tasks and presented the first approach for certified segmentation. 
To perform image segmentation, each pixel in an image is assigned a segmentation class. This can be seen as a type of classification task, but instead of predicting the content of the entire image, the goal is to predict the class of each individual pixel.
In this setting, the target space corresponds to regions/categories to segment (\eg, cars, roads, pedestrians, etc.), and the classifier $f: \Rbb^{d} \rightarrow \Ycal^d$ outputs a class for each pixel and classifies each component individually.
It is relatively straightforward to extend the certification algorithm proposed by~\cite{cohen2019certified} for the segmentation task.
Nevertheless, \cite{fischer2021scalable} identified two primary challenges with the method.
First, given that the certified radius of a particular region will be the minimum radius over the entire region, the algorithm may report an extremely low certified radius based on only a few {\em bad} pixels.
Second, since  \citeauthor{cohen2019certified}'s certification algorithm is applied to each pixel separately, and the certification is only valid with a probability of $1 - \alpha$, considering the entire region and applying the union bound could significantly reduce the overall confidence in the certificate.
To address the first challenge and limit the impact on bad pixels on the overall result, \cite{fischer2021scalable} proposed a simple solution which consists in defining a threshold $\tau \in [\frac12, 1]$ and instead of checking $\underline{p_y} > \frac12$, they advise  $\underline{p_y} > \tau$.
To account for the multiple testing problem, \ie, low confidence due to the union bound of the entire region, \cite{fischer2021scalable} introduce the $\texttt{FwerControl}$ function used in Algorithm~\ref{alg:segcertify} which is based on the Holm-Bonferroni method~\citep{holm1979simple}, and performs multiple-testing correction.
Conceptually, the idea is to control the type I error (rejecting the null hypothesis when it is actually true) while reducing type II errors (not rejecting the null hypothesis when it is false).
Now that we have reviewed randomized smoothing for classification and segmentation tasks, we will present how it is possible to improve upon the current state-of-the-art with diffusion models.

\begin{table*}[ht]
  \centering
  \caption{ Segmentation results of \textsc{DenoiseCertify} (ours) and \textsc{SegCertify} proposed by \cite{fischer2021scalable} on both Cityscapes and Pascal-Context datasets. Two network architectures were used for both pipelines, HRNet trained with noise and ViT trained without noise. We also report the performance of HRNet trained without noise. For each dataset, we used the same 100 images with $n_0 = 10,  n=100, \alpha = 0.001$ and $\tau = 0.75$. Results are certified at radius $R$, acc. being the mean per-pixel accuracy, mIoU the mean intersection over union and \%\abstain the mean percentage of pixel abstentions on all images.}
  \resizebox{\textwidth}{!}{
  \begin{tabular}{lllccccccccccc}
  \toprule
    \multirow{2}{*}{\textbf{Model}} & & \multirow{2}{*}{\textbf{Architecture}} & \multirow{2}{*}{\shortstack{\textbf{Trained} \\ \textbf{with noise}}} & \multirow{2}{*}{\boldsymbol{$\sigma$}} & \multirow{2}{*}{\boldsymbol{$R$}}
    & \multicolumn{3}{c}{\textbf{Cityscapes}} & & \multicolumn{3}{c}{\textbf{Pascal Context}} \\
    \cmidrule{7-9} \cmidrule{11-13}
    & & & & & & Acc & mIoU & \%\abstain && Acc. &  mIoU & \%\abstain \\
    \midrule
    \multirow{1}{*}{\shortstack[l]{Non-robust}}
    & & HRNet & \xmark & 0.00 & 0.00 & 0.97 & 0.81 & 0.00 & & 0.77 &  0.42 & 0.00 \\
    \midrule
    \multirow{4}{*}{\shortstack[l]{\textsc{SegCertify}}}
    & & \multirow{4}{*}{HRNet} & \cmark & 0.00 & 0.00 & 0.91  & 0.57 & 0.00 && 0.53 &  0.18 & 0.00 \\
    & & & \cmark & 0.25 & 0.17 & 0.88 & 0.59 & 0.11 && 0.55 & 0.22 & 0.22 \\
    & & & \cmark & 0.50 & 0.34 & 0.34 & 0.06 & 0.40 && 0.17 & 0.03 & 0.41 \\
    & & & \cmark & 1.00 & 0.67 & 0.06 & 0.00 & 0.31 && 0.08 & 0.00 & 0.13 \\
    \midrule
    \multirow{7}{*}{\shortstack[l]{\textsc{DenoiseCertify} \\ \textbf{(ours)}}}
    & & \multirow{3}{*}{\shortstack[l]{Diffusion\\+HRNet}}
    & \cmark & 0.25 & 0.17 & 0.70 & 0.32 & 0.26 && 0.47 & 0.17 & 0.27 \\
    & & & \cmark & 0.50 & 0.34 & 0.55 & 0.21 & 0.41 && 0.42 & 0.15 & 0.46 \\
    & & & \cmark & 1.00 & 0.67 & 0.36 & 0.09 & 0.60 && 0.15 & 0.04 & 0.77 \\
    \cmidrule{3-13}
    & & \multirow{4}{*}{\shortstack[l]{Diffusion\\+ViT}} & \xmark & 0.00 & 0.00 & 0.94 & 0.67 & 0.00 && 0.85 &  0.58 & 0.00 \\
    & & & \xmark & 0.25 & 0.17 & 0.77 & 0.41 & 0.24 && 0.67 & 0.48 & 0.28 \\
    & & & \xmark & 0.50 & 0.34 & 0.65 & 0.28 & 0.36 && 0.54 & 0.32 & 0.40 \\
    & & & \xmark & 1.00 & 0.67 & 0.47 & 0.15 & 0.53 && 0.28 & 0.15 & 0.62 \\
  \bottomrule
  \end{tabular}
  \label{table:result1}
  }
\end{table*}

\section{Certified Segmentation via Diffusion Models}

To prevent a distribution shift when using randomized smoothing for inference, it is common practice to train networks with noise injection~\citep{cohen2019certified}.
However, from an information theory perspective, randomized smoothing has inherent trade-offs and limitations. 
While adding noise during training can enhance the certified accuracy of models compared to those trained without noise, it may also lower the model's natural accuracy, as the variance of the noise decreases the information present in the input. 
These limitations have led to a series of no-go results for randomized smoothing~\citep{blum2020random,hayes2020extensions,kumar2020curse,yang2020randomized,mohapatra2021hidden,wu2021completing,ettedgui2022towards}, suggesting that achieving high certified accuracy may be challenging due to the significant variance that must be introduced in the input.
Consequently, the destruction of information due to noise can result in information loss, potentially leading to a useless classifier.

To address this important limitation of randomized smoothing, ~\cite{salman2020denoised} have investigated denoising the input before giving it to the classifier.
The idea is to use trained neural networks to \emph{reconstruct} the removed information of the image due to the noise.
This process has two main advantages: it mitigates the no-go results of randomized smoothing since the destroyed information is ``reconstructed'' by the denoiser and it does not involve training the classifier with noise mitigating the reduced natural accuracy of training with noise injection.
Of course, in this new setting, the quality of the denoiser will matter.
\cite{salman2020denoised} were able to boost the certified accuracy by up to 33\% on the ImageNet dataset with respect to previous state-of-the-art defenses.

\cite{carlini2023certified} go even further and propose to use state-of-the-art Diffusion Probabilistic Models (DPM) to perform the denoising step.
With this approach, they were able to further improve the state of the art by up to 14\% on the ImageNet dataset.
Denoising diffusion probabilistic models, which have been introduced by~\cite{sohl2015deep} and further improved by~\cite{ho2020denoising} and \cite{nichol2021improved}, are a class of generative models and have shown to beat Generative Adversarial Networks (GANs) \citep{goodfellow2020generative} on image synthesis.
Conceptually, the training of these models consists in adding noise at each step of the diffusion process until purely random noise is reached.
The reverse process then starts from random noise and generates a new image from the data distribution.
\cite{carlini2023certified} proposes a procedure to use these models for {\em denoising} instead of generating new images.
The idea is to start the reverse process with a noisy image instead of Gaussian noise in order for the DPM to output an image from the initial data distribution that resembles the original image.
As explained by~\cite{carlini2023certified}, to use the DPM in the context of randomized smoothing, one needs to convert the noise added for randomized smoothing, \ie, $x_\text{rs} = x + \tau$ with $\tau \sim \Ncal(0, \sigma^2 \Imat)$ to the specific step in the diffusion process: $x_\text{DPM} = \sqrt{\beta_t} x + \tau (1 - \beta_t)$ where $\beta_t$ denotes a constant from the timestamp $t$ that controls the amount of noise added to the image during the diffusion process. 
For more details on how to compute the timestamp $t$, one can refer to Section 3 of \cite{carlini2023certified}.
We provide in Algorithm~\ref{algo:denoise} an updated version of the algorithm to compute the samples for the Predict \& Certify function of \cite{fischer2021scalable}.

\paragraph{Pipeline.}
Our pipeline starts by passing the image through a Denoising Diffusion Probabilistic Model (DDPM) and then calling a semantic segmentation model for prediction. 
For both components, we use an off-the-shelf model made publicly available. To denoise images, we use the class unconditional DDPM from~\cite{dhariwal2021diffusion}. This 552M-parameter denoiser has been trained on ImageNet and performs very well on images from both Cityscapes and Pascal-Context. For segmentation, we use two model architectures with different training strategies. First, we test on High-resolution networks, HRNet from \cite{wang2020deep}, trained in two different ways. The \textit{non-robust} HRNet has been trained with natural images and the \textit{base model} is an HRNet trained with a Gaussian noise of $\sigma = 0.25$.
The second architecture we use is the Vision Transformer Adapter for Dense Predictions, ViT from \cite{chen2023vision}, that was trained only on natural images. We use the 568M-parameter model trained on Pascal-Context and the 571M-parameter model trained on Cityscapes. Both models were reported to provide state-of-the-art accuracy and mean intersection over union (mIoU) on the task of semantic segmentation. Our code is provided at: \url{https://github.com/othmanela/certified_segmentation}

\begin{table*}[ht]
  \centering
  \caption{Performance of \textsc{SegCertify} and \textsc{DenoiseCertify} (ours) on an off-the-shelf HRNet model trained without Gaussian noise. Scale corresponds to the image sizing scale used as input to the segmentation model (\eg, a scale of 0.5 on cityscapes will resize the images to $512 \times 1024$). Accuracy, mean intersection over union (mIoU) and percentage of abstentions (\%\abstain) are certified given a noise level $\sigma$ and radius $R$. All results are provided with Holm correction.}
  \label{table:result2}
  \begin{tabular}{cllccccccccccc}
  \toprule
  \multirow{2}{*}{\textbf{Scale}} & & \multirow{2}{*}{\textbf{Model}} & \multirow{2}{*}{\boldsymbol{$\sigma$}} & \multirow{2}{*}{\boldsymbol{$R$}} & & \multicolumn{3}{c}{\textbf{Cityscapes}} & & \multicolumn{3}{c}{\textbf{Pascal Context}} \\
  \cmidrule{7-9} \cmidrule{11-13}
  & & & & & & Acc. & mIoU & \%\abstain & & Acc. & mIoU & \%\abstain \\
  \midrule
  \multirow{6}{*}{0.25} & \phantom{} &
        \multirow{3}{*}{\textsc{SegCertify}}
            & 0.25 & 0.17 & & 0.34 & 0.05 & 0.29 & & 0.18 & 0.05 & 0.67 \\
        & & & 0.50 & 0.34 & & 0.18 & 0.01 & 0.14 & & 0.07 & 0.01 & 0.70 \\
        & & & 1.00 & 0.67 & & 0.18 & 0.01 & 0.06 & & 0.02 & 0.00 & 0.58 \\
        \cmidrule{3-13}
        & & \multirow{3}{*}{\textsc{DenoiseCertify}}
            & 0.25 & 0.17 & & 0.78 & 0.41 & 0.22 & & 0.45 & 0.19 & 0.28 \\
        & & & 0.50 & 0.34 & & 0.68 & 0.29 & 0.32 & & 0.37 & 0.11 & 0.45 \\
        & & & 1.00 & 0.67 & & 0.46 & 0.15 & 0.54 & & 0.11 & 0.03 & 0.74 \\
  \midrule
  \multirow{6}{*}{0.50} & &
        \multirow{3}{*}{\textsc{SegCertify}}
            & 0.25 & 0.17 & & 0.48 & 0.07 & 0.19 & & 0.34 & 0.13 & 0.49 \\
        & & & 0.50 & 0.34 & & 0.19 & 0.01 & 0.15 & & 0.12 & 0.03 & 0.62 \\
        & & & 1.00 & 0.67 & & 0.17 & 0.01 & 0.11 & & 0.03 & 0.00 & 0.43 \\
        \cmidrule{3-13}
        & & \multirow{3}{*}{\textsc{DenoiseCertify}}
            & 0.25 & 0.17 & & 0.74 & 0.37 & 0.26 & & 0.56 & 0.23 & 0.27 \\
        & & & 0.50 & 0.34 & & 0.60 & 0.22 & 0.40 & & 0.43 & 0.18 & 0.46 \\
        & & & 1.00 & 0.67 & & 0.32 & 0.11 & 0.67 & & 0.12 & 0.04 & 0.64 \\
  \midrule
  \multirow{6}[2]{*}{1.00} & &
        \multirow{3}{*}{\textsc{SegCertify}}
            & 0.25 & 0.17 & & 0.18 & 0.01 & 0.08 & & 0.31 & 0.10 & 0.52 \\
        & & & 0.50 & 0.34 & & 0.17 & 0.01 & 0.08 & & 0.08 & 0.01 & 0.45 \\
        & & & 1.00 & 0.67 & & 0.01 & 0.00 & 0.98 & & 0.01 & 0.02 & 0.49 \\
        \cmidrule{3-13}
        & & \multirow{3}{*}{\textsc{DenoiseCertify}}
            & 0.25 & 0.17 & & 0.58 & 0.26 & 0.42 & & 0.47 & 0.22 & 0.27 \\
        & & & 0.50 & 0.34 & & 0.42 & 0.15 & 0.55 & & 0.36 & 0.10 & 0.53 \\
        & & & 1.00 & 0.67 & & 0.24 & 0.70 & 0.70 & & 0.10 & 0.02 & 0.73 \\
  \bottomrule
  \end{tabular}%
\end{table*}%

\begin{table*}[ht]
  \centering
  \caption{Comparison of two denoising strategies on the \textsc{DenoiseCertify} pipeline. All of the reported results use a Vision Transformer (ViT) segmentation model with a scale of $1$. For the denoising diffusion model, we use the same timestep $t^*$ found with the \textsc{computeTimestep} function presented in Algorithm ~\ref{algo:denoise}.}
  \label{table:result4}
  \begin{tabular}{lcccrrrrrrrrr}
  \toprule
    \multirow{2}{*}{\textbf{Denoise Method}} & \multirow{2}{*}{\boldsymbol{$\sigma$}} & \multirow{2}{*}{\boldsymbol{$R$}} &  & \multicolumn{3}{c}{\textbf{Cityscapes}} &  & \multicolumn{3}{c}{\textbf{Pascal Context}} \\
    \cmidrule{5-7}
    \cmidrule{9-12}
    & & & & Acc. &  mIoU & \%\abstain & & Acc. &  mIoU & \%\abstain \\
    \midrule
    \multirow{4}{*}{Denoise single step}
    & 0.00 & 0.00 & & 0.94 & 0.67 & 0.00 & & 0.85 & 0.58 & 0.00 \\
    & 0.25 & 0.17 & & 0.77 & 0.41 & 0.24 & & 0.67 & 0.48 & 0.28 \\
    & 0.50 & 0.34 & & 0.65 & 0.28 & 0.36 & & 0.54 & 0.32 & 0.40 \\
    & 1.00 & 0.67 & & 0.47 & 0.15 & 0.53 & & 0.28 & 0.15 & 0.62 \\
    \midrule
    \multirow{4}{*}{Denoise multi-step}
    & 0.00 & 0.00 & & 0.89 & 0.39 & 0.00 & & 0.80 & 0.49 & 0.00 & \\
    & 0.25 & 0.17 & & 0.70 & 0.27 & 0.33 & & 0.61 & 0.34 & 0.34 & \\
    & 0.50 & 0.34 & & 0.52 & 0.15 & 0.50 & & 0.46 & 0.24 & 0.49 & \\
    & 1.00 & 0.67 & & 0.29 & 0.06 & 0.73 & & 0.14 & 0.07 & 0.75 & \\
  \bottomrule
  \end{tabular}
\end{table*}

\section{Experiments}\label{section:experiements}

We evaluate our method on a set of experiments with multiple approaches.
First, we compare our technique with \textsc{SegCertify}, the state-of-the-art introduced by \cite{fischer2021scalable}.
Then, we set new state-of-the-art results using off-the-shelf models. We name our method \textsc{DenoiseCertify}. 

\paragraph{Datasets.}
All of our experiments are performed on the task of semantic image segmentation on Pascal-Context and Cityscapes datasets, two very common datasets for this task. 
Pascal-Context~\citep{mottaghi2014role} consists of an extension of the Pascal-VOC~\citep{pascalvoc} dataset with all of the image pixels annotated. There are 60 classes (59 foreground and 1 background). Typical evaluation strategies use either all of the 60 classes or the 59 foreground classes only. We evaluate here on the 59 foreground classes in order to have a fair comparison with \textsc{SegCertify}. The Cityscapes dataset ~\citep{cordts2016cityscapes} contains high resolution $1024 \times 2048$ images of diverse street scenes from 50 different cities. The images are annotated in $30$ classes but only $19$ of them are used for evaluation.
Similar to \textsc{SegCertify}, we evaluate our method on the same $100$ images set from both datasets. We use every 5\textsuperscript{th} image on the Cityscapes dataset and every 51\textsuperscript{st} on Pascal.

\paragraph{\textsc{DenoiseCertify} on models trained with noise.}
We start first by comparing \textsc{DenoiseCertify} with \textsc{SegCertify}. The state-of-the-art certification results proposed by the latter were obtained with an HRNet trained with a Gaussian noise of $\sigma = 0.25$. A comparison of both methods is provided in the first two sections of Table ~\ref{table:result1}. We notice that \textsc{DenoiseCertify} outperforms \textsc{SegCertify} for  all sigmas except for $0.25$. In fact, for $ \sigma = 0.5$ the accuracy jumps from 0.34 to 0.55 and the mIoU from 0.06 to 0.21 which corresponds to an increase of 61\% and 250\% respectively. This gives us an idea of the power of denoising diffusion models when used to certify segmentation models. 
Since our pipeline contains an added denoising step, we note an increase in the reported runtime in seconds. On the largest images of the dataset ($1024 \times 2048$), the runtime increases from $92.69$ to $131.42$ seconds with HRNet, which per image is minor. We did not perform any optimization on the code to make our pipeline faster. With more engineering, the runtime can be optimized further. Also, we believe that the gain in performance easily outweighs the increase in runtime. For $ \sigma = 1.0$, it appears from Table ~\ref{table:result1} that \textsc{SegCertify} has a lower number of abstentions than \textsc{DenoiseCertify}. However, looking at the segmentations it looks like \textsc{SegCertify} predicts a large number of pixels in the image with the wrong class. An example is provided in the last row of Figure \ref{figure:cityscapes}. For $ \sigma = 0.25$, \textsc{SegCertify} outperforms our technique but this may be due to two main reasons.
First, the model we are using was trained with a Gaussian noise of $0.25$. Thus, it is performing  best when provided with images with the same level of noise. In the next section, we show that given the right model, \textsc{DenoiseCertify} outperforms \textsc{SegCertify}.
Second, one of the limitations of using the off-the-shelf denoiser provided by \cite{dhariwal2021diffusion} is rescaling the images to $256 \times 256$. Therefore scaling them back to their original size may decrease the image quality, especially when using high-resolution images like Cityscapes. We perform experiments with multiple scales and report them in the subsequent section.

\begin{figure}[t]
  \centering
  \hfill
  \begin{subfigure}[h]{0.23\textwidth}
    \centering
    \includegraphics[width=\textwidth]{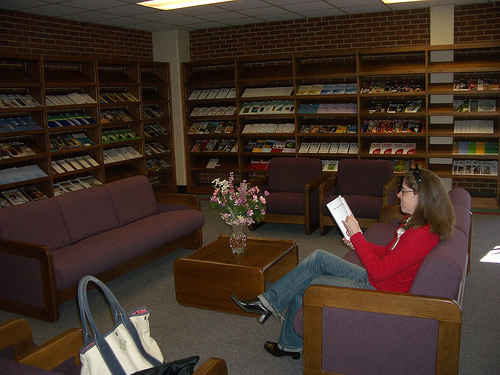} \\[0.2cm]
    \includegraphics[width=\textwidth]{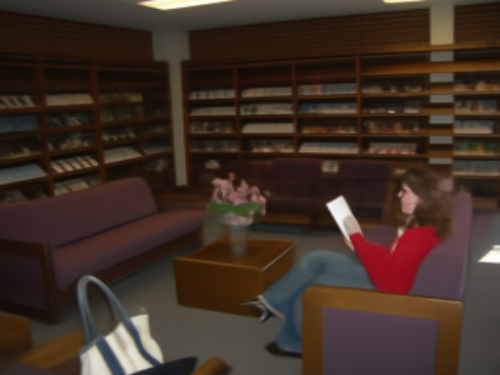}
  \end{subfigure}
  \hfill
  \begin{subfigure}[h]{0.23\textwidth}
    \centering
    \includegraphics[width=\textwidth]{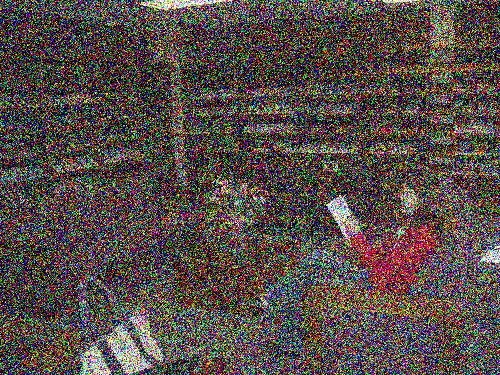} \\[0.2cm]
    \includegraphics[width=\textwidth]{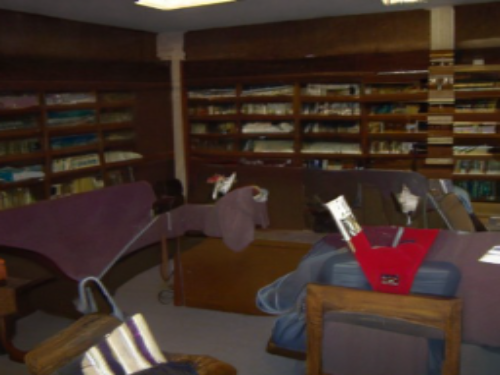}
  \end{subfigure}
  \caption{Qualitative results of the performance of a denoising diffusion model on Pascal Context images. Top row from left to right: ground truth and ground truth with a noise of $\sigma = 1.0$. Bottom row: single-step denoised image and multistep denoised image. }
  \label{figure:ddpm}
\end{figure}

\paragraph{\textsc{DenoiseCertify} on non-robust models.}
Here we use an HRNet model that was trained on natural images only without any introduction of Gaussian noise. We compare our performance with \textsc{SegCertify} and report our results in Table \ref{table:result2}. Focusing on a scale of 1, we notice that \textsc{SegCertify} achieves a poor performance. In fact, as $\sigma$ increases to values $> 0.25$ the mIoU and accuracy end up becoming 0. This is an expected result since models trained on natural images are very sensitive to Gaussian noise. However, those types of models perfectly suit our methodology, as we denoise images, we are able to use off-the-shelf segmentation models and achieve a much better prediction. 
This introduces a paradigm shift as we no longer require training robust deep learning models that need highly engineered strategies and that also degrade the natural accuracy significantly. As reported in Table \ref{table:result1}, when comparing the first two rows, the non-robust HRNet accuracy drops from 0.97 to 0.91 and the decrease is more significant for the mIoU, going from 0.81 to 0.57. Therefore, our technique allows us to limit the drop in performance that traditional models used to suffer from while keeping strong certification guarantees.

\begin{figure}[t]
  \centering
  \hfill
  \begin{subfigure}[h]{0.18\textwidth}
    \centering
    \includegraphics[width=\textwidth]{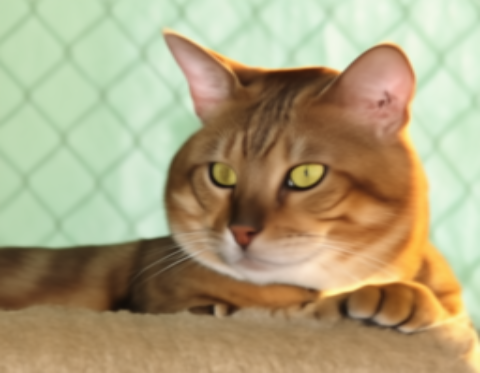} 
  \end{subfigure}
  \hfill
  \begin{subfigure}[h]{0.28\textwidth}
    \centering
    \includegraphics[width=\textwidth]{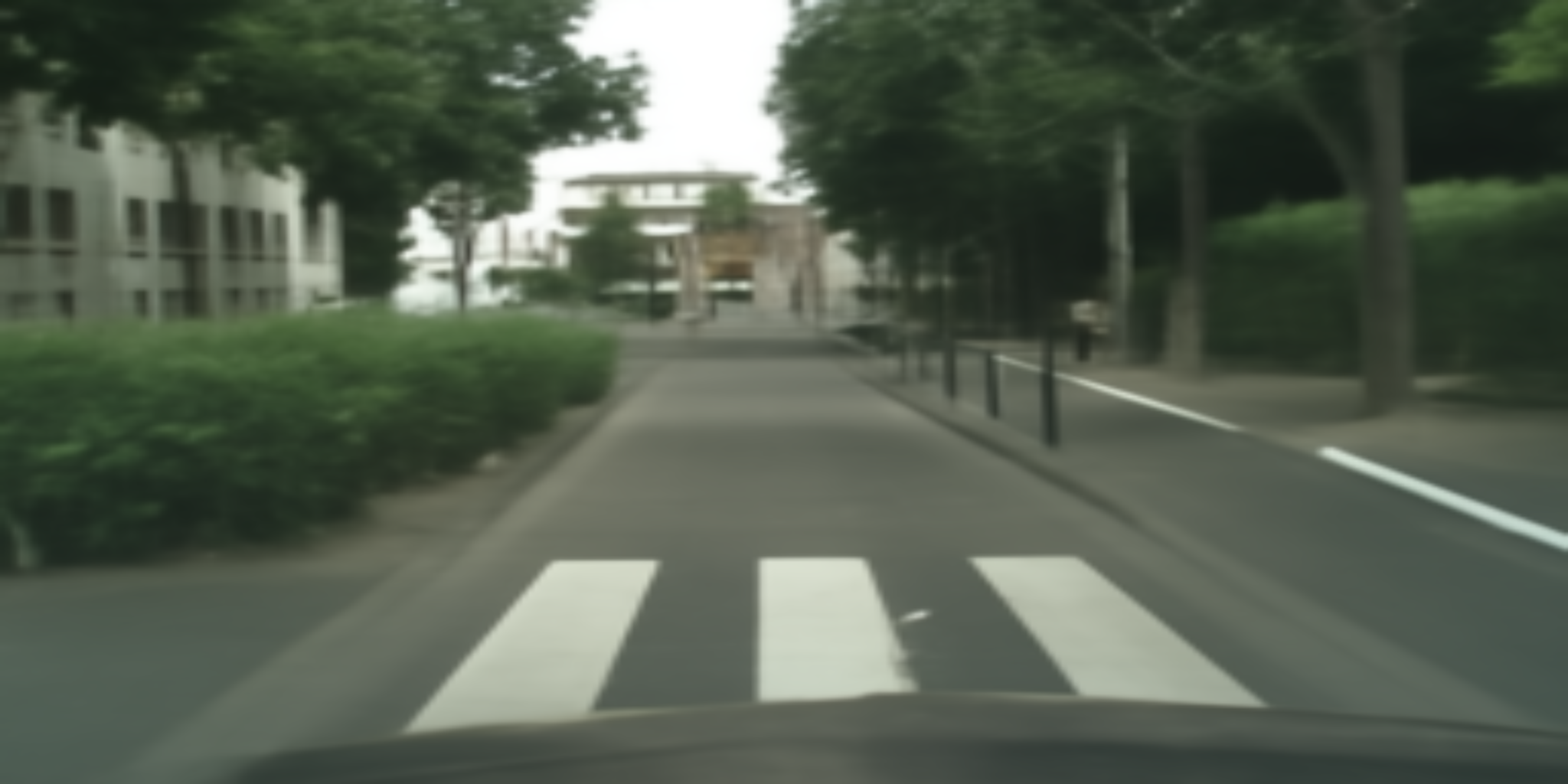}
  \end{subfigure}
  \caption{Examples of denoised and upscaled images from the Denoising Diffusion Model. On the left, a Pascal-Context image denoised on size $256 \times 256$ and upscaled to $373 \times 480$. On the right, a Cityscapes image denoised on $256 \times 256$ and upscaled to $1024 \times 2048$.}
  \label{figure:upscale}
\end{figure}

\paragraph{Impact of image scales on performance.}
One of the limitations of using off-the-shelf models is having to comply with their restrictions. The unconditional DDPM we are using only takes as input images of size $256 \times 256$. We thus have to downscale the input images and upscale them back to their original size for prediction. As stated above, this is the main limitation of our method. But, since semantic segmentation models can be invoked with multiple scales, we can use them to predict at a given scale and then upsample the output probabilities back to the original size of the image. This also has the advantage of providing faster predictions. As an example, at a scale of 0.5 for Cityscapes, we downsample the images to $256 \times 256$ in order to call the DDPM, the denoised image is then reshaped to $512 \times 1024$ and serves as input to the segmentation model. The output probabilities of the segmentation model are then upsampled to their original size ($1024 \times 2048$) to be compared with the ground truth.
We always perform the certification on the original size of the image in order to follow the same strategy as \textsc{SegCertify} and perform a fair comparison.
The performance of both methods with multiple scales is reported in Table \ref{table:result2}. Examples of denoised and upscaled images are provided in Figure \ref{figure:upscale}. For Cityscapes, we notice that the smaller the scale, the better the performance. In fact, the accuracy jumps from 0.58 to 0.74, and 0.78 for scale values of 1, 0.5, and 0.25 respectively. 
\textsc{DenoiseCertify} performs best for a scale of 0.25, corresponding to a Cityscapes image size of $256 \times 512$ which is very close to the output of the DDPM. Therefore, the rescaling does not impact the details and overall quality of the denoised image.
The same happens for the Pascal-Context dataset, the best performance is obtained for a scale of 0.5 which corresponds to images of size $240 \times 240$, again very close to the $256 \times 256$ DDPM output.
Training DDPMs on images of higher resolution would be another way to circumvent this limitation.
Also, with the improvement of powerful techniques that rely on the denoising backbone, our approach would still be able to leverage the resources made available. We believe that having a denoiser that is also able to upscale images to very high resolutions would allow us to improve our results even further.

\paragraph{\textsc{DenoiseCertify} on state-of-the-art segmentation models.}
So far we have discussed how \textsc{DenoiseCertify} performs on both a robust and non-robust HRNet model. We have empirically shown that it achieves the best results on standard deep learning-based segmentation models.
Going a step further, we can leverage the power of Vision Transformers which have been reported to be more robust to attacks ~\citep{Mao_2022_CVPR}, but also give state-of-the-art results on semantic segmentation tasks \citep{chen2023vision}.
In this section, we use the ViT Adapter trained with natural images and report the results in the last section of Table \ref{table:result1}. When comparing with the results of Table \ref{table:result2} on the same scale of 1, we notice that the ViT model provides a considerable increase. For the lowest $\sigma = 0.25$, the accuracy and mIoU are respectively boosted to 0.77 and 0.41 compared to 0.58 and 0.26 previously. This empirically proves the points of ~\cite{Mao_2022_CVPR} and would even encourage us to make the assumption that we would be able to obtain higher certification results with stronger transformer models. 
Overall, \textsc{DenoiseCertify} combined with a ViT achieves state-of-the-art semantic segmentation certification results for Pascal-Context and Cityscapes. 
 
\paragraph{Generalization of Denoising Diffusion Models.}
As stated above, we use an off-the-shelf denoising diffusion model that was trained on ImageNet. 
We set the number of channels to $256$ and apply a linear scheduler with $1000$ steps. Qualitative results are provided in Figure \ref{figure:ddpm}. We clearly notice that the DDPM is able the denoise the image with the highest level of noise ($\sigma = 1.0$) while keeping all of the information of the picture. Therefore, it is important to state that diffusion models generalize well to datasets they were not trained on. Pascal-Context and Cityscapes are the first two examples. Future work will involve testing DDPMs on images from other distributions (\eg, the medical domain).

\paragraph{Multistep Denoising Diffusion Model.}
Denoising diffusion probabilistic models have been introduced as a class of generative models that beat GANs on image synthesis~\citep{dhariwal2021diffusion}. Starting with Gaussian noise, each step of the DDPM consists in denoising an input image at timestep $t$ to a marginally less noisy image at timestep $t-1$. The complete diffusion process is an iterative procedure starting from $t^*$ until $t = 0$.
Programmatically, each call to the denoiser $d$ at timestep $t$ performs two actions; it predicts the completely denoised image and returns the average between the estimated denoised image and the noisy image of timestep $t-1$. 
We conduct experiments on the two possible denoising strategies. The top section of Table \ref{table:result4} reports the results of a single-step denoised image prediction from the class unconditional DDPM. The bottom section of Table \ref{table:result4} on the other hand reports results of a multiple-step denoising strategy going from $t^*$ until $t = 0$ iteratively on the same class unconditional DDPM. Both use the ViT as the segmentation model.
From the presented results, it is clear that the single-step denoiser performs better than the multi-step one in terms of accuracy, mIoU, percentage of abstentions, and runtime. This shows that denoising the image in a single shot is better than repeatedly denoising it multiple times. Intuitively, since the DDPMs are generative models at heart, they will tend to behave as such when denoising an image multiple times. Therefore, the output image at $t = 0$ may have lost a lot of its original information or may even end up from a different distribution. Qualitative results from Figure \ref{figure:ddpm} support this claim as we can clearly notice that elements of the image were removed in the multi-step approach (the flower pot, as well as the reader disappeared, and the shape of the furniture changed). Another advantage of single-step denoising is the runtime efficiency. Instead of having to call the denoiser multiple times passing the outputted image at each timestep $t$, the denoiser is only called once (\eg, For $\sigma=1.0$ a denoiser with linear scheduling will be called 258 times compared to a single time with the first scheme). This represents a nonnegligible advantage of the single-shot denoising since we are using multiple calls to the denoiser for each image in order to obtain the certificate.
We deduce that denoising diffusion models are powerful but should be used accordingly. In our case, we would like to leverage the denoising properties of the DDPM more than their generative properties. Thus, a single-step denoising strategy should be adopted.

\section{Conclusion}

We present the first work on certified semantic segmentation that leverages denoising diffusion probabilistic models and vision transformers. 
We conduct a comprehensive set of experiments on Pascal-Context~\citep{mottaghi2014role} and Cityscapes~\citep{cordts2016cityscapes} datasets and show that our method achieves state-of-the-art results on certified robustness for semantic segmentation tasks. 
We were able to achieve significant improvements in accuracy and mIoU using off-the-shelf models that are not trained or fine-tuned for robustness. 
This work provides a new direction for certified image segmentation with {\em off-the-shelf} models. However, an interesting direction would be to explore task-specific training. For instance, in the context of certified segmentation, \cite{salman2019provably} improved upon the work of~\cite{cohen2019certified} by training classifiers with noise injection and adversarial training. It would be straightforward to extend this approach to certified segmentation with our \textsc{DenoiseCertify} procedure by adversarially training a classifier or a diffusion model. Although computationally expensive, this method may lead to further improvements. 
Moreover, we have seen that the diffusion model is able to generalize to Pascal-Context and Cityscapes datasets.
A promising future direction would be to investigate the generalization of this model for denoising medical images and provide certified segmentation for critical healthcare applications.

\begin{acknowledgements} % will be removed in pdf for initial submission,
% (without ‘accepted’ option in \documentclass)
% so you can already fill it to test with the
% ‘accepted’ class option
This work was granted access to the HPC resources of IDRIS under the allocation 2023-AD011013308R1 made by GENCI.
\end{acknowledgements}

\bibliography{bibliography}

\begin{thebibliography}{58}
\providecommand{\natexlab}[1]{#1}
\providecommand{\url}[1]{\texttt{#1}}
\expandafter\ifx\csname urlstyle\endcsname\relax
  \providecommand{\doi}[1]{doi: #1}\else
  \providecommand{\doi}{doi: \begingroup \urlstyle{rm}\Url}\fi

\bibitem[Araujo et~al.(2020)Araujo, Meunier, Pinot, and
  Negrevergne]{araujo2020advocating}
Alexandre Araujo, Laurent Meunier, Rafael Pinot, and Benjamin Negrevergne.
\newblock Advocating for multiple defense strategies against adversarial
  examples.
\newblock In \emph{ECML PKDD 2020 Workshops 2020}, pages 165--177. Springer,
  2020.

\bibitem[Araujo et~al.(2021)Araujo, Negrevergne, Chevaleyre, and
  Atif]{araujo2021lipschitz}
Alexandre Araujo, Benjamin Negrevergne, Yann Chevaleyre, and Jamal Atif.
\newblock On lipschitz regularization of convolutional layers using toeplitz
  matrix theory.
\newblock \emph{Proceedings of the AAAI Conference on Artificial Intelligence},
  pages 6661--6669, 2021.

\bibitem[Araujo et~al.(2023)Araujo, Havens, Delattre, Allauzen, and
  Hu]{araujo2023a}
Alexandre Araujo, Aaron~J Havens, Blaise Delattre, Alexandre Allauzen, and Bin
  Hu.
\newblock A unified algebraic perspective on lipschitz neural networks.
\newblock In \emph{The Eleventh International Conference on Learning
  Representations}, 2023.

\bibitem[Arnab et~al.(2018)Arnab, Miksik, and Torr]{arnab2018robustness}
Anurag Arnab, Ondrej Miksik, and Philip~HS Torr.
\newblock On the robustness of semantic segmentation models to adversarial
  attacks.
\newblock In \emph{Proceedings of the IEEE Conference on Computer Vision and
  Pattern Recognition}, pages 888--897, 2018.

\bibitem[Athalye et~al.(2018)Athalye, Carlini, and
  Wagner]{athalye2018obfuscated}
Anish Athalye, Nicholas Carlini, and David Wagner.
\newblock Obfuscated gradients give a false sense of security: Circumventing
  defenses to adversarial examples.
\newblock In \emph{International conference on machine learning}, pages
  274--283. PMLR, 2018.

\bibitem[Blum et~al.(2020)Blum, Dick, Manoj, and Zhang]{blum2020random}
Avrim Blum, Travis Dick, Naren Manoj, and Hongyang Zhang.
\newblock Random smoothing might be unable to certify $\ell_\infty$ robustness
  for high-dimensional images.
\newblock \emph{The Journal of Machine Learning Research}, 21\penalty0
  (1):\penalty0 8726--8746, 2020.

\bibitem[Carlini and Wagner(2017)]{carlini2017towards}
Nicholas Carlini and David Wagner.
\newblock Towards evaluating the robustness of neural networks.
\newblock In \emph{2017 ieee symposium on security and privacy (sp)}, pages
  39--57. Ieee, 2017.

\bibitem[Carlini et~al.(2023)Carlini, Tramer, Dvijotham, Rice, Sun, and
  Kolter]{carlini2023certified}
Nicholas Carlini, Florian Tramer, Krishnamurthy~Dj Dvijotham, Leslie Rice,
  Mingjie Sun, and J~Zico Kolter.
\newblock (certified!!) adversarial robustness for free!
\newblock In \emph{The Eleventh International Conference on Learning
  Representations}, 2023.

\bibitem[Chen et~al.(2023)Chen, Duan, Wang, He, Lu, Dai, and
  Qiao]{chen2023vision}
Zhe Chen, Yuchen Duan, Wenhai Wang, Junjun He, Tong Lu, Jifeng Dai, and
  Yu~Qiao.
\newblock Vision transformer adapter for dense predictions.
\newblock In \emph{The Eleventh International Conference on Learning
  Representations}, 2023.

\bibitem[Cohen et~al.(2019)Cohen, Rosenfeld, and Kolter]{cohen2019certified}
Jeremy Cohen, Elan Rosenfeld, and Zico Kolter.
\newblock Certified adversarial robustness via randomized smoothing.
\newblock In \emph{international conference on machine learning}, pages
  1310--1320. PMLR, 2019.

\bibitem[Cordts et~al.(2016)Cordts, Omran, Ramos, Rehfeld, Enzweiler, Benenson,
  Franke, Roth, and Schiele]{cordts2016cityscapes}
Marius Cordts, Mohamed Omran, Sebastian Ramos, Timo Rehfeld, Markus Enzweiler,
  Rodrigo Benenson, Uwe Franke, Stefan Roth, and Bernt Schiele.
\newblock The cityscapes dataset for semantic urban scene understanding.
\newblock In \emph{Proceedings of the IEEE conference on computer vision and
  pattern recognition}, pages 3213--3223, 2016.

\bibitem[Croce and Hein(2020)]{croce2020reliable}
Francesco Croce and Matthias Hein.
\newblock Reliable evaluation of adversarial robustness with an ensemble of
  diverse parameter-free attacks.
\newblock In \emph{International conference on machine learning}, pages
  2206--2216. PMLR, 2020.

\bibitem[Croce and Hein(2021)]{croce2021mind}
Francesco Croce and Matthias Hein.
\newblock Mind the box: $ l\_1 $-apgd for sparse adversarial attacks on image
  classifiers.
\newblock In \emph{International Conference on Machine Learning}, pages
  2201--2211. PMLR, 2021.

\bibitem[Dhariwal and Nichol(2021)]{dhariwal2021diffusion}
Prafulla Dhariwal and Alexander Nichol.
\newblock Diffusion models beat gans on image synthesis.
\newblock \emph{Advances in Neural Information Processing Systems},
  34:\penalty0 8780--8794, 2021.

\bibitem[Ettedgui et~al.(2022)Ettedgui, Araujo, Pinot, Chevaleyre, and
  Atif]{ettedgui2022towards}
Raphael Ettedgui, Alexandre Araujo, Rafael Pinot, Yann Chevaleyre, and Jamal
  Atif.
\newblock Towards evading the limits of randomized smoothing: A theoretical
  analysis.
\newblock \emph{arXiv preprint arXiv:2206.01715}, 2022.

\bibitem[Everingham et~al.(2015)Everingham, Eslami, Van~Gool, Williams, Winn,
  and Zisserman]{pascalvoc}
M.~Everingham, S.~M.~A. Eslami, L.~Van~Gool, C.~K.~I. Williams, J.~Winn, and
  A.~Zisserman.
\newblock The pascal visual object classes challenge: A retrospective.
\newblock \emph{International Journal of Computer Vision}, 111, 2015.

\bibitem[Farnia et~al.(2019)Farnia, Zhang, and Tse]{farnia2018generalizable}
Farzan Farnia, Jesse Zhang, and David Tse.
\newblock Generalizable adversarial training via spectral normalization.
\newblock In \emph{International Conference on Learning Representations}, 2019.

\bibitem[Fischer et~al.(2021)Fischer, Baader, and Vechev]{fischer2021scalable}
Marc Fischer, Maximilian Baader, and Martin Vechev.
\newblock Scalable certified segmentation via randomized smoothing.
\newblock In \emph{International Conference on Machine Learning}, pages
  3340--3351. PMLR, 2021.

\bibitem[Goodfellow et~al.(2020)Goodfellow, Pouget-Abadie, Mirza, Xu,
  Warde-Farley, Ozair, Courville, and Bengio]{goodfellow2020generative}
Ian Goodfellow, Jean Pouget-Abadie, Mehdi Mirza, Bing Xu, David Warde-Farley,
  Sherjil Ozair, Aaron Courville, and Yoshua Bengio.
\newblock Generative adversarial networks.
\newblock \emph{Communications of the ACM}, 63\penalty0 (11):\penalty0
  139--144, 2020.

\bibitem[Goodfellow et~al.(2014)Goodfellow, Shlens, and
  Szegedy]{goodfellow2014explaining}
Ian~J Goodfellow, Jonathon Shlens, and Christian Szegedy.
\newblock Explaining and harnessing adversarial examples.
\newblock \emph{arXiv preprint arXiv:1412.6572}, 2014.

\bibitem[Gowal et~al.(2018)Gowal, Dvijotham, Stanforth, Bunel, Qin, Uesato,
  Arandjelovic, Mann, and Kohli]{gowal2018effectiveness}
Sven Gowal, Krishnamurthy Dvijotham, Robert Stanforth, Rudy Bunel, Chongli Qin,
  Jonathan Uesato, Relja Arandjelovic, Timothy Mann, and Pushmeet Kohli.
\newblock On the effectiveness of interval bound propagation for training
  verifiably robust models.
\newblock \emph{arXiv preprint arXiv:1810.12715}, 2018.

\bibitem[Hayes(2020)]{hayes2020extensions}
Jamie Hayes.
\newblock Extensions and limitations of randomized smoothing for robustness
  guarantees.
\newblock In \emph{Proceedings of the IEEE/CVF Conference on Computer Vision
  and Pattern Recognition Workshops}, pages 786--787, 2020.

\bibitem[He et~al.(2019)He, Yang, Li, Li, Chang, and Yu]{he2019non}
Xiang He, Sibei Yang, Guanbin Li, Haofeng Li, Huiyou Chang, and Yizhou Yu.
\newblock Non-local context encoder: Robust biomedical image segmentation
  against adversarial attacks.
\newblock \emph{Proceedings of the AAAI Conference on Artificial Intelligence},
  2019.

\bibitem[Ho et~al.(2020)Ho, Jain, and Abbeel]{ho2020denoising}
Jonathan Ho, Ajay Jain, and Pieter Abbeel.
\newblock Denoising diffusion probabilistic models.
\newblock \emph{Advances in Neural Information Processing Systems},
  33:\penalty0 6840--6851, 2020.

\bibitem[Holm(1979)]{holm1979simple}
Sture Holm.
\newblock A simple sequentially rejective multiple test procedure.
\newblock \emph{Scandinavian journal of statistics}, pages 65--70, 1979.

\bibitem[Huang et~al.(2021)Huang, Zhang, Shi, Kolter, and
  Anandkumar]{huang2021training}
Yujia Huang, Huan Zhang, Yuanyuan Shi, J~Zico Kolter, and Anima Anandkumar.
\newblock Training certifiably robust neural networks with efficient local
  lipschitz bounds.
\newblock \emph{Advances in Neural Information Processing Systems},
  34:\penalty0 22745--22757, 2021.

\bibitem[Kang et~al.(2020)Kang, Song, Du, and Guizani]{kang2020adversarial}
Xu~Kang, Bin Song, Xiaojiang Du, and Mohsen Guizani.
\newblock Adversarial attacks for image segmentation on multiple lightweight
  models.
\newblock \emph{IEEE Access}, 8:\penalty0 31359--31370, 2020.

\bibitem[Kumar and Goldstein(2021)]{kumar2021center}
Aounon Kumar and Tom Goldstein.
\newblock Center smoothing: Certified robustness for networks with structured
  outputs.
\newblock \emph{Advances in Neural Information Processing Systems},
  34:\penalty0 5560--5575, 2021.

\bibitem[Kumar et~al.(2020)Kumar, Levine, Goldstein, and Feizi]{kumar2020curse}
Aounon Kumar, Alexander Levine, Tom Goldstein, and Soheil Feizi.
\newblock Curse of dimensionality on randomized smoothing for certifiable
  robustness.
\newblock In \emph{International Conference on Machine Learning}, pages
  5458--5467. PMLR, 2020.

\bibitem[Kurakin et~al.(2018)Kurakin, Goodfellow, and
  Bengio]{kurakin2018adversarial}
Alexey Kurakin, Ian~J Goodfellow, and Samy Bengio.
\newblock Adversarial examples in the physical world.
\newblock In \emph{Artificial intelligence safety and security}, pages 99--112.
  Chapman and Hall/CRC, 2018.

\bibitem[Lecuyer et~al.(2019)Lecuyer, Atlidakis, Geambasu, Hsu, and
  Jana]{lecuyer2019certified}
Mathias Lecuyer, Vaggelis Atlidakis, Roxana Geambasu, Daniel Hsu, and Suman
  Jana.
\newblock Certified robustness to adversarial examples with differential
  privacy.
\newblock In \emph{2019 IEEE Symposium on Security and Privacy (SP)}, pages
  656--672. IEEE, 2019.

\bibitem[Li et~al.(2019{\natexlab{a}})Li, Chen, Wang, and
  Carin]{li2019certified}
Bai Li, Changyou Chen, Wenlin Wang, and Lawrence Carin.
\newblock Certified adversarial robustness with additive noise.
\newblock \emph{Advances in neural information processing systems}, 32,
  2019{\natexlab{a}}.

\bibitem[Li et~al.(2019{\natexlab{b}})Li, Haque, Anil, Lucas, Grosse, and
  Jacobsen]{li2019preventing}
Qiyang Li, Saminul Haque, Cem Anil, James Lucas, Roger~B Grosse, and
  J{\"o}rn-Henrik Jacobsen.
\newblock Preventing gradient attenuation in lipschitz constrained
  convolutional networks.
\newblock \emph{Advances in neural information processing systems}, 32,
  2019{\natexlab{b}}.

\bibitem[Madry et~al.(2018)Madry, Makelov, Schmidt, Tsipras, and
  Vladu]{madry2017towards}
Aleksander Madry, Aleksandar Makelov, Ludwig Schmidt, Dimitris Tsipras, and
  Adrian Vladu.
\newblock Towards deep learning models resistant to adversarial attacks.
\newblock In \emph{International Conference on Learning Representations}, 2018.

\bibitem[Mao et~al.(2022)Mao, Qi, Chen, Li, Duan, Ye, He, and
  Xue]{Mao_2022_CVPR}
Xiaofeng Mao, Gege Qi, Yuefeng Chen, Xiaodan Li, Ranjie Duan, Shaokai Ye, Yuan
  He, and Hui Xue.
\newblock Towards robust vision transformer.
\newblock In \emph{Proceedings of the IEEE/CVF Conference on Computer Vision
  and Pattern Recognition (CVPR)}, pages 12042--12051, June 2022.

\bibitem[Meunier et~al.(2022)Meunier, Delattre, Araujo, and
  Allauzen]{meunier2022dynamical}
Laurent Meunier, Blaise~J Delattre, Alexandre Araujo, and Alexandre Allauzen.
\newblock A dynamical system perspective for lipschitz neural networks.
\newblock In \emph{International Conference on Machine Learning}, pages
  15484--15500. PMLR, 2022.

\bibitem[Miyato et~al.(2018)Miyato, Kataoka, Koyama, and
  Yoshida]{miyato2018spectral}
Takeru Miyato, Toshiki Kataoka, Masanori Koyama, and Yuichi Yoshida.
\newblock Spectral normalization for generative adversarial networks.
\newblock In \emph{International Conference on Learning Representations}, 2018.

\bibitem[Mohapatra et~al.(2020)Mohapatra, Ko, Weng, Chen, Liu, and
  Daniel]{mohapatra2020higher}
Jeet Mohapatra, Ching-Yun Ko, Tsui-Wei Weng, Pin-Yu Chen, Sijia Liu, and Luca
  Daniel.
\newblock Higher-order certification for randomized smoothing.
\newblock \emph{Advances in Neural Information Processing Systems},
  33:\penalty0 4501--4511, 2020.

\bibitem[Mohapatra et~al.(2021)Mohapatra, Ko, Weng, Chen, Liu, and
  Daniel]{mohapatra2021hidden}
Jeet Mohapatra, Ching-Yun Ko, Lily Weng, Pin-Yu Chen, Sijia Liu, and Luca
  Daniel.
\newblock Hidden cost of randomized smoothing.
\newblock In \emph{International Conference on Artificial Intelligence and
  Statistics}, pages 4033--4041. PMLR, 2021.

\bibitem[Mottaghi et~al.(2014)Mottaghi, Chen, Liu, Cho, Lee, Fidler, Urtasun,
  and Yuille]{mottaghi2014role}
Roozbeh Mottaghi, Xianjie Chen, Xiaobai Liu, Nam-Gyu Cho, Seong-Whan Lee, Sanja
  Fidler, Raquel Urtasun, and Alan Yuille.
\newblock The role of context for object detection and semantic segmentation in
  the wild.
\newblock In \emph{Proceedings of the IEEE conference on computer vision and
  pattern recognition}, pages 891--898, 2014.

\bibitem[Nichol and Dhariwal(2021)]{nichol2021improved}
Alexander~Quinn Nichol and Prafulla Dhariwal.
\newblock Improved denoising diffusion probabilistic models.
\newblock In \emph{International Conference on Machine Learning}, pages
  8162--8171. PMLR, 2021.

\bibitem[Pinot et~al.(2019)Pinot, Meunier, Araujo, Kashima, Yger, Gouy-Pailler,
  and Atif]{pinot2019theoretical}
Rafael Pinot, Laurent Meunier, Alexandre Araujo, Hisashi Kashima, Florian Yger,
  C{\'e}dric Gouy-Pailler, and Jamal Atif.
\newblock Theoretical evidence for adversarial robustness through
  randomization.
\newblock \emph{Advances in neural information processing systems}, 32, 2019.

\bibitem[Prach and Lampert(2022)]{prach2022almost}
Bernd Prach and Christoph~H Lampert.
\newblock Almost-orthogonal layers for efficient general-purpose lipschitz
  networks.
\newblock In \emph{Computer Vision--ECCV 2022: 17th European Conference, Tel
  Aviv, Israel, October 23--27, 2022, Proceedings, Part XXI}, pages 350--365.
  Springer, 2022.

\bibitem[Salman et~al.(2019)Salman, Li, Razenshteyn, Zhang, Zhang, Bubeck, and
  Yang]{salman2019provably}
Hadi Salman, Jerry Li, Ilya Razenshteyn, Pengchuan Zhang, Huan Zhang, Sebastien
  Bubeck, and Greg Yang.
\newblock Provably robust deep learning via adversarially trained smoothed
  classifiers.
\newblock \emph{Advances in Neural Information Processing Systems}, 32, 2019.

\bibitem[Salman et~al.(2020)Salman, Sun, Yang, Kapoor, and
  Kolter]{salman2020denoised}
Hadi Salman, Mingjie Sun, Greg Yang, Ashish Kapoor, and J~Zico Kolter.
\newblock Denoised smoothing: A provable defense for pretrained classifiers.
\newblock \emph{Advances in Neural Information Processing Systems},
  33:\penalty0 21945--21957, 2020.

\bibitem[Singla and Feizi(2021)]{singla2021skew}
Sahil Singla and Soheil Feizi.
\newblock Skew orthogonal convolutions.
\newblock In \emph{International Conference on Machine Learning}, pages
  9756--9766. PMLR, 2021.

\bibitem[Sohl-Dickstein et~al.(2015)Sohl-Dickstein, Weiss, Maheswaranathan, and
  Ganguli]{sohl2015deep}
Jascha Sohl-Dickstein, Eric Weiss, Niru Maheswaranathan, and Surya Ganguli.
\newblock Deep unsupervised learning using nonequilibrium thermodynamics.
\newblock In \emph{International Conference on Machine Learning}, pages
  2256--2265. PMLR, 2015.

\bibitem[Szegedy et~al.(2013)Szegedy, Zaremba, Sutskever, Bruna, Erhan,
  Goodfellow, and Fergus]{szegedy2013intriguing}
Christian Szegedy, Wojciech Zaremba, Ilya Sutskever, Joan Bruna, Dumitru Erhan,
  Ian Goodfellow, and Rob Fergus.
\newblock Intriguing properties of neural networks.
\newblock \emph{arXiv preprint arXiv:1312.6199}, 2013.

\bibitem[Trockman and Kolter(2021)]{trockman2021orthogonalizing}
Asher Trockman and J~Zico Kolter.
\newblock Orthogonalizing convolutional layers with the cayley transform.
\newblock In \emph{International Conference on Learning Representations}, 2021.

\bibitem[Tsuzuku et~al.(2018)Tsuzuku, Sato, and Sugiyama]{tsuzuku2018lipschitz}
Yusuke Tsuzuku, Issei Sato, and Masashi Sugiyama.
\newblock Lipschitz-margin training: Scalable certification of perturbation
  invariance for deep neural networks.
\newblock \emph{Advances in neural information processing systems}, 31, 2018.

\bibitem[Wang et~al.(2020)Wang, Sun, Cheng, Jiang, Deng, Zhao, Liu, Mu, Tan,
  Wang, et~al.]{wang2020deep}
Jingdong Wang, Ke~Sun, Tianheng Cheng, Borui Jiang, Chaorui Deng, Yang Zhao,
  Dong Liu, Yadong Mu, Mingkui Tan, Xinggang Wang, et~al.
\newblock Deep high-resolution representation learning for visual recognition.
\newblock \emph{IEEE transactions on pattern analysis and machine
  intelligence}, 43\penalty0 (10):\penalty0 3349--3364, 2020.

\bibitem[Wu et~al.(2021)Wu, Bojchevski, Kuvshinov, and
  G{\"u}nnemann]{wu2021completing}
Yihan Wu, Aleksandar Bojchevski, Aleksei Kuvshinov, and Stephan G{\"u}nnemann.
\newblock Completing the picture: Randomized smoothing suffers from the curse
  of dimensionality for a large family of distributions.
\newblock In \emph{International Conference on Artificial Intelligence and
  Statistics}, pages 3763--3771. PMLR, 2021.

\bibitem[Xiang et~al.(2019)Xiang, Qi, and Li]{xiang2019generating}
Chong Xiang, Charles~R Qi, and Bo~Li.
\newblock Generating 3d adversarial point clouds.
\newblock In \emph{Proceedings of the IEEE/CVF Conference on Computer Vision
  and Pattern Recognition}, pages 9136--9144, 2019.

\bibitem[Xie et~al.(2017)Xie, Wang, Zhang, Zhou, Xie, and
  Yuille]{xie2017adversarial}
Cihang Xie, Jianyu Wang, Zhishuai Zhang, Yuyin Zhou, Lingxi Xie, and Alan
  Yuille.
\newblock Adversarial examples for semantic segmentation and object detection.
\newblock In \emph{Proceedings of the IEEE international conference on computer
  vision}, pages 1369--1378, 2017.

\bibitem[Xu et~al.(2022)Xu, Li, and Li]{xu2022lot}
Xiaojun Xu, Linyi Li, and Bo~Li.
\newblock {LOT}: Layer-wise orthogonal training on improving l2 certified
  robustness.
\newblock In Alice~H. Oh, Alekh Agarwal, Danielle Belgrave, and Kyunghyun Cho,
  editors, \emph{Advances in Neural Information Processing Systems}, 2022.

\bibitem[Yang et~al.(2020)Yang, Duan, Hu, Salman, Razenshteyn, and
  Li]{yang2020randomized}
Greg Yang, Tony Duan, J~Edward Hu, Hadi Salman, Ilya Razenshteyn, and Jerry Li.
\newblock Randomized smoothing of all shapes and sizes.
\newblock In \emph{International Conference on Machine Learning}, pages
  10693--10705. PMLR, 2020.

\bibitem[Yatsura et~al.(2023)Yatsura, Sakmann, Hua, Hein, and
  Metzen]{yatsura2022certified}
Maksym Yatsura, Kaspar Sakmann, N.~Grace Hua, Matthias Hein, and Jan~Hendrik
  Metzen.
\newblock Certified defences against adversarial patch attacks on semantic
  segmentation.
\newblock In \emph{The Eleventh International Conference on Learning
  Representations}, 2023.

\bibitem[Yu et~al.(2022)Yu, Li, Cai, and Li]{yu2022constructing}
Tan Yu, Jun Li, Yunfeng Cai, and Ping Li.
\newblock Constructing orthogonal convolutions in an explicit manner.
\newblock In \emph{International Conference on Learning Representations}, 2022.

\end{thebibliography}
\end{document}